\documentclass[preprints,article,accept,moreauthors,pdftex]{Definitions/mdpi} 
\firstpage{1} 
\pubvolume{1}
\issuenum{1}
\articlenumber{0}
\pubyear{2022}
\copyrightyear{2022}
\datereceived{} 
\dateaccepted{} 
\datepublished{} 
\hreflink{https://doi.org/} 

\usepackage{algorithm}
\usepackage[noend]{algpseudocode}
\usepackage{caption}
\usepackage{subcaption}

\Title{Improved LiDAR Odometry and Mapping using Deep Semantic Segmentation and Novel Outliers Detection}

\TitleCitation{Improved LiDAR Odometry and Mapping using Deep Semantic Segmentation and Novel Outliers Detection}

\Author{Mohamed Afifi $^{1,*}$\orcidA{} and Mohamed ElHelw $^{1}$}

\AuthorNames{Mohamed Afifi and Mohamed ElHelw}

\AuthorCitation{Afifi, M; ElHelw, M.}

\address[1]{%
$^{1}$ \quad Center for Informatics Science, Nile University, Giza, Egypt}

\corres{Correspondence: moh.afifi@nu.edu.eg}

\abstract{Perception is a key element for enabling intelligent autonomous navigation. Understanding the semantics of the surrounding environment and accurate vehicle pose estimation are essential capabilities for autonomous vehicles, including self-driving cars and mobile robots that perform complex tasks. Fast moving platforms like self-driving cars impose a hard challenge for localization and mapping algorithms. In this work, we propose a novel framework for real-time LiDAR odometry and mapping based on LOAM architecture for fast moving platforms. Our framework utilizes semantic information produced by a deep learning model to improve point-to-line and point-to-plane matching between LiDAR scans and build a semantic map of the environment, leading to more accurate motion estimation using LiDAR data. We observe that including semantic information in the matching process introduces a new type of outlier matches to the process, where matching occur between different objects of the same semantic class. To this end, we propose a novel algorithm that explicitly identifies and discards potential outliers in the matching process. In our experiments, we study the effect of improving the matching process on the robustness of LiDAR odometry against high speed motion. Our experimental evaluations on KITTI dataset demonstrate that utilizing semantic information and rejecting outliers significantly enhance the robustness of LiDAR odometry and mapping when there are large gaps between scan acquisition poses, which is typical for fast moving platforms.}

\keyword{LiDAR Odometry; Semantic Segmentation; Outliers Rejection} 

\begin{document}




\section{Introduction}
On-board real time localization and mapping is essential for mobile robots that operate in unknown environments. The ability of robots to build geometric semantically labelled maps can enable intelligent autonomous navigation in unexplored places. Highly dynamic environments usually pose additional challenges to the process of reliable localization and mapping. 

One of the most commonly used sensors in localization and mapping is the Light Detection and Ranging (LiDAR) sensor. 3D LiDAR sensors consist of one or more laser beams that are steered by the sensor driver to scan the 3D volume surrounding the sensor. Geometrical relationships between two consecutive LiDAR scans can be used to estimate the vehicle motion between the two scans, which is known as odometry estimation. LiDAR odometry aims to estimate the motion of the sensor between two consecutive scans using point cloud registration, which is the estimation of the transformation that should be applied to the points of one scan to get the two scans to overlap as good as possible. Most state of the art LiDAR odometry techniques like \cite{zhang2017low} start by matching points from one scan to lines and planes in the other scan, then use optimization techniques to estimate the motion transformation that minimizes the distance between these points and their corresponding lines and planes. The matching process is usually simple and is based on simple nearest neighbors search. Therefore, points from one scan and their matches in the other scan may not correspond to the same object in space, especially if the initial alignment between the two scans is relatively far away from the true alignment, which will cause some of the matches to be false matches. Fast motion of LiDAR makes motion estimation using scan matching even more challenging. This is because most approaches use simple velocity models to provide a rough initial guess for the motion between the two scans, which is refined using scan matching. When the LiDAR scanning frequency is low relative to the vehicle speed, the initial alignment will no longer be good enough to provide good matches using naive nearest neighbors search. LiDAR odometry algorithms are very sensitive to such outliers in the matching process, and therefore robust regression techniques \cite{fox2002robust} are usually used when estimating the motion transformation. Another source of error is moving objects, which can introduce more outliers to the process even if the initial alignment between the scans is relatively good. Some recent work try to detect and remove points that correspond to potentially moving objects to make the matching process more robust \cite{deschaud2018imls, moreno2020study}. 

Recent advances in deep learning made it possible to perform real-time semantic segmentation of point clouds generated from LiDAR scans, which is the process of assigning a semantic label (e.g. car, pedestrian, building, ...) to every point in the LiDAR scan. There are publicly available datasets (e.g. \cite{behley2019semantickitti}) that can be used to train deep learning models to perform end-to-end semantic segmentation of 3D point clouds. Exploiting semantic information has high potential for improving the accuracy of localization and mapping, because knowing the semantic type of each point can improve the matching process between point clouds. It can also be used to filter out potentially moving objects, which can make the motion estimation more robust and construct the map using only static objects.

The objectives of this work can be summarized as follows:

\begin{itemize}
    \item Propose a framework for robust semantic localization and mapping that fully integrates semantic information with LOAM \cite{zhang2017low} architecture.
    \item Demonstrate that exploiting semantic information in the matching process can lead to outlier matches between different objects of the same class, and introduce a novel outlier detection and rejection technique to further refine the selected matches.
    \item Study the impact of improving matches using semantic information and outliers rejection on the robustness of scan registration and motion estimation for fast moving vehicles.
\end{itemize}

The rest of the paper is organized as follows: Section \ref{sec:related_work} reviews the related work, Section \ref{sec:proposed_framework} describes the proposed framework, Section \ref{sec:results} discusses the experimental evaluation of our approach, and in Section \ref{sec:future} we conclude our work and state future work directions.

\section{Related Work} \label{sec:related_work}

Many approaches have been proposed for registering two point clouds. The most popular family of methods is the Iterative Closest Point (ICP) algorithm \cite{121791} and its variants. The basic version of the ICP algorithm iterates between assigning point-to-point matches using nearest neighbors search and estimating the transformation that minimizes the mean of squared distances between these matches until the process converges. In the case of point clouds acquired from LiDAR scans, the estimated transformation that registers the two point clouds represents the relative pose transformation between the LiDAR poses at the time of acquisition of the two scans. Instead of matching points to points, other ICP variants were proposed that match points to local surface patches. For example, \cite{132043} proposed point-to-plane matching. The current state of the art framework for LiDAR odometry and mapping is known as LOAM \cite{zhang2017low}. The authors of LOAM use point-to-line and point-to-plane ICP to estimate the motion between consecutive LiDAR scans and update the current pose estimation. 

Pose estimation using scan-to-scan LiDAR odometry (registration of two consecutive scans) introduces small errors at each time step. Such errors accumulate over time and cause the estimated pose to drift away from the ground truth pose. To reduce this drift, most approaches perform scan-to-map localization (registration of the current scan to a map) to refine the pose estimated from the scan-to-scan odometry. The map is usually a point cloud aggregated over previous time steps. Registration of the current scan to a map collected over multiple previous scans gives more accurate pose estimation results than registration to just the previous scan. Implicit Surface Least Squares (IMLS) SLAM \cite{deschaud2018imls} registers the current scan to a map aggregated over the previous $n$ time steps, and report the results for $n$ = 1, 5, 10, 100. Instead of using raw points during registration, IMLS transforms the map into an implicit surface representation \cite{curless1996volumetric}. LOAM \cite{zhang2017low} on the other hand aggregates the map point cloud over all previous time steps. To prevent the number of points in the map from getting too large, they use voxel-grid downsampling to discretize the map into voxels and keep only one point inside each voxel (the centroid of all points inside that voxel). To further reduce the drift caused by accumulating errors, many approaches (e.g. \cite{behley2018efficient, chen2019suma++}) use loop closure. Loop closure process starts by loop detection, which detects when the vehicle re-visits a place it has visited before. Once the loop is detected and verified, the error between the estimated current pose and the pose from the previous visit is propagated to correct all poses in the trajectory between the two visits, and consequently correct the constructed map.

Exploiting semantic information in LiDAR localization and mapping has become an active area of research. SegMap \cite{doi:10.1177/0278364919863090} extracts segments from the point cloud using clustering, then semantically classifies each segment into one of three classes (vehicle, build, other). Instead of low level iterative closest point scan matching, SegMap adopts a segment to segment matching framework. They state that rejecting segments from potentially dynamic objects increase the robustness of the localization algorithm. More recently, SUMA++ \cite{chen2019iros} starts by projecting the 3D point cloud on a spherical surface, then uses RangeNet++ \cite{milioto2019rangenet++} to segment this spherical projection, followed by motion estimation between the two consequtive LiDAR scans directly from their semantically segmented spherical projections. They use semantic information to weight the residuals (perpendicular distance between a point and its matching plane or line) according to the semantic compatibility between matches, such that costs corresponding to matches with consistent semantic labels are given higher weight in the optimization process according to the probabilities that RangeNet++ assigns to the semantic labels. They finally build a semantic map of the environment using surfel-based representation, and use a spherical image view of the surfel-based map rendered using the LiDAR pose at each time step to perform scan-to-map registration to further refine the estimated pose. In \cite{8387438}, the authors use PointNet \cite{qi2017pointnet} to provide semantic segmentations for the points clouds, and exploit this semantic information to select matches based on object types. However, their approach is limited to scan-to-scan registration using generalized ICP \cite{segal2009generalized} without generating semantic LiDAR maps, and therefore lacks the opportunity to refine the estimated pose using scan-to-map registration. The authors of \cite{8387438} also proposed an older version \cite{zaganidis2017semantic} that uses Normal Distribution Transform \cite{biber2003normal} for point cloud registration instead of generalized ICP. \cite{parkison2018semantic} perform joint semantic and geometric inference using Expectation-Maximization. Their approach however is very slow because they do not perform explicit point-to-point matching, but rather assign weights to all possible available matches and take semantic information into consideration when assigning those weights. Their approach also does not exploit scan-to-map pose refinement.

Many approaches try to improve the performance of LOAM \cite{zhang2017low} framework by exploiting semantic segmentations of the LiDAR scans. \cite{deschaud2018imls, moreno2020study} showed that filtering out points that belong to potentially dynamic object classes improves the LOAM scan matching process. \cite{9012188} exploits the probability vector over semantic classes that his neural network assigns to each point in the LiDAR scan and uses it during the least-squares optimization to improve the motion transformation estimation by weighting the costs (squared errors) of all matches according to the similarity between the probability vector of each point and its match. \cite{du2021lidar} uses the ground-truth semantic labels provided by SemanticKITTI dataset \cite{behley2019semantickitti} to improve the module that extracts feature points that lie on edges and planar surfaces and to filter out dynamic objects. They also include semantic constraints in the equations that compute the matching costs to use only matches between correspondences that have the same semantic label. Unlike methods that utilize semantic information in the cost computation step of the optimization process, our framework constructs $\mathbb{C}$ separate KDTrees \cite{bentley1975multidimensional} from the input point cloud, one KDTree for each semantic class, where $\mathbb{C}$ is the total number of classes that the semantic segmentation model assigns to different points in different scans. Then search for matches between two scans only in the KDTree of the corresponding semantic label. This has the advantage of increasing the probability of finding correct matches even if the distance between a point and its true match is larger than the distance to false matches from different classes as we explain in Section \ref{sec:semantic_odometry}, while maintaining fast processing speed of the framework.

LeGO-LOAM \cite{shan2018lego} performs ground plane segmentation, then tries to find matches between points that lie on planar surfaces in the current scan only with points that lie on the ground plane in the previous scan, and search for matches between points that lie on edges in the current scan with non-ground points in the previous scan. Their approach however is limited to ground segmentation, and doesn't perform full semantic segmentation using deep learning to detect different semantic classes and exploit them in the matching process. \cite{8949363} proposes a framework designed particularly for LiDAR odometry and mapping in forests, they semantically classify points in the point cloud into tree and ground points. They use point-to-plane for matching ground points and point-to-cylinder for matching tree points. 


Most SLAM approaches use M-estimation techniques \cite{fox2002robust} (e.g. Huber estimator) for robust motion estimation in the presence of outliers. M-estimation is equivalent to assigning weights to each residual in the least-squares problem according to its cost. The cost in this context is the squared distance between the point and the line or plane to which it was matched. Residuals that correspond to large distances are treated as potential outliers and are assigned lower weights to reduce their influence on the estimated transformation. The downside of this approach is that outlier matches that we get from nearest neighbors search can still correspond to a small cost, which means that outliers may be assigned high weights and can highly influence the optimization process. In our experiments, we show that our proposed outliers rejection technique further improves the robustness of the framework even when M-estimation techniques are employed, as it explicitly identifies and reject outlier matches.

\section{Proposed Framework}  \label{sec:proposed_framework}
\begin{figure*}[htp]
    \centering
	\includegraphics[width=0.8\textwidth, keepaspectratio]{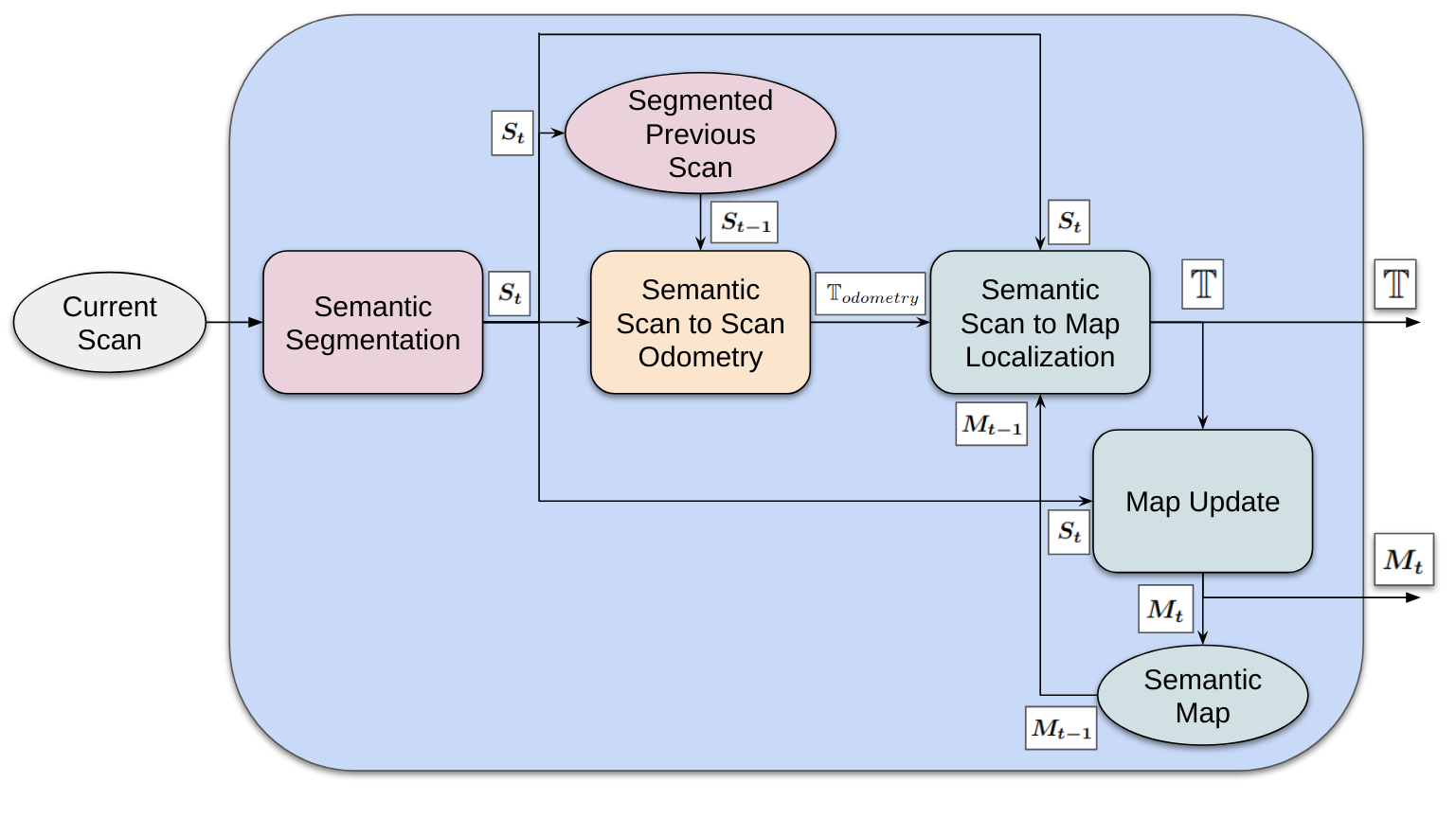}
	\vspace{3pt}
	\caption{Block diagram of our proposed framework.}
	\label{fig:block_diagram}
\end{figure*}

In this section, we discuss the details of our proposed framework for LiDAR odometry and mapping. The proposed framework is based on the LOAM architecture \cite{zhang2017low} the state of the art LiDAR odometry method on KITTI benchmark \cite{geiger2012we}. We further improve the basic LOAM architecture by exploiting semantic information from a deep semantic segmentation model, and discard outliers from the matching process using a novel outliers detection and rejection algorithm. A block diagram of our framework can be found in Figure \ref{fig:block_diagram}.

\subsection{Deep Semantic Segmentation}
We start by passing the current LiDAR scan through a real-time deep semantic segmentation network. Our deep learning model assigns a semantic class label to each point in the scan. The semantic segmentation module runs on a separate thread in parallel to the rest of the framework to enhance computational efficiency.

We developed a Multi-Projection Fusion (MPF) pipeline for semantic segmentation of 3D point clouds that relies on using two 2D projections of the point cloud where each projection is processed by an independent CNN model. The pipeline starts by feeding the input point cloud to two branches, one responsible for spherical view projection and the other for bird's-eye view projection. Each branch applies semantic segmentation on the projected point cloud. Subsequently, predictions from the two branches are fused to produce the final prediction. In our previous work \cite{DBLP:journals/corr/abs-2011-01974} we experimentally show that the fusion of two views improves the segmentation accuracy over using each view independently. More information about the used CNN architectures and training setup can be found at our paper \cite{DBLP:journals/corr/abs-2011-01974}.

\subsection{Semantic LiDAR Odometry} \label{sec:semantic_odometry}
Similar to LOAM \cite{zhang2017low}, our framework starts by extracting keypoints on planar and edge surfaces from the current scan and match them to planes and edges from the previous scan, while taking the semantic information into account to yield better matches. We adopt an ICP variant that estimates the motion between two scans by minimizing the mean of squared distances between point-to-line and point-to-plane matches. The estimated motion is used to further refine the selected keypoint matches using our novel outliers rejection algorithm. The process of refining matches and re-estimating the motion using the new set of matches is repeated till convergence. The pseudocode in Algorithm \ref{alg:sem_lidar_odom} summarizes the main steps of our semantic LiDAR odometry. The odometry transformation $\mathbb{T}_{odometry}$ transforms points from the coordinate frame defined by the LiDAR pose at the current time step to the coordinate frame defined by the LiDAR pose at the previous time step. $\mathbb{T}_{odometry}$ is initialized using the motion transformation estimated from the previous time step, then refined using the procedure described in Algorithm \ref{alg:sem_lidar_odom} by estimating the transformation that registers the semantically-segmented current and previous point clouds $\boldsymbol{S}_t$ and $\boldsymbol{S}_{t-1}$.

\begin{algorithm}[tb]

\footnotesize
    \begin{algorithmic}[]
    \State \textbf{Initialize} $\mathbb{T}_{odometry}$
    \State \textbf{Initialize} $iter \gets 1$\\
    
    \While {no convergence of $\mathbb{T}_{odometry}$ and $iter \leq iters_{max}$}
        \State // Extract keypoints from the current and previous clouds 
        \State $\mathbb{S}_t \gets ExtractKeypoints(\boldsymbol{S}_t)$ 
        \State $\mathbb{S}_{t-1} \gets ExtractKeypoints(\boldsymbol{S}_{t-1})$ \\
        
        \State // Assign matches 
        \State $\mathbb{M} \gets \emptyset$ 
        \For{${}^t\boldsymbol{p_{curr}} \in \mathbb{S}_t$}
            \State $\boldsymbol{{}^{t-1}p_{curr}} \gets \mathbb{T}_{odometry} * {}^t\boldsymbol{p_{curr}}$ 
            \State $\mathbb{N} \gets SemanticNearestNeighbors(\boldsymbol{{}^{t-1}p_{curr}}, \mathbb{S}_{t-1})$ 
            \texttt{\\}
            
            \State // Fit a plane or line (depending on the surface type of $\boldsymbol{p_{curr}}$) and add the match to the set $\mathbb{M}$ 
            \State $planeOrLine \gets FitPlaneOrLine(\mathbb{N})$
            \State $\mathbb{M} \gets \mathbb{M}  \cup  \{({}^t\boldsymbol{p_{curr}}, planeOrLine)\}$
        \EndFor
        \texttt{\\}
        
        \State // ORME stands for Outliers Rejection and Motion Estimation in Algorithm \ref{alg:orme} 
        \State $\mathbb{T}_{odometry} \gets ORME(\mathbb{T}_{odometry}, \mathbb{M})$ 
        
        \State $iter$ += 1 
    \EndWhile

    \texttt{\\}
    \State return $\mathbb{T}_{odometry}$
    \end{algorithmic}
    
\caption{SemanticLidarOdometry($\boldsymbol{S}_t, \boldsymbol{S}_{t-1}$)}
\label{alg:sem_lidar_odom}

\end{algorithm}

\subsubsection{Notation}
We denote the coordinates of a point $\boldsymbol{p}$ with respect to an arbitrary Cartesian coordinate frame $\boldsymbol{F}$ as $\boldsymbol{{}^Fp}$ ($\boldsymbol{p}$ preceded by a superscript $\boldsymbol{F}$). We define the transformation that transforms the coordinates of a point $\boldsymbol{p}$ from a frame $\boldsymbol{A}$ to frame $\boldsymbol{B}$ as a pair $\mathbb{T} = (\boldsymbol{t}, \boldsymbol{q})$, where $\boldsymbol{t}$ is a 3D translation vector and $\boldsymbol{q}$ is a 4D unit quaternion \cite{schmidt2001using} corresponding to the rotation part of the transformation. We use the operator $*$ to apply the rotation and translation implied by the transformation as follows:

\begin{equation}
\label{eq:apply_transformation}
    \boldsymbol{{}^Bp} = \mathbb{T} * \boldsymbol{{}^Ap}
    = \boldsymbol{q}.\boldsymbol{{}^Ap}.\boldsymbol{q}^{-1} + \boldsymbol{t}
\end{equation}

Where $\boldsymbol{{}^Ap}$ and $\boldsymbol{{}^Bp}$ are the coordinates of the 3D point $\boldsymbol{p}$ relative to coordinate frames $\boldsymbol{A}$ and $\boldsymbol{B}$, respectively.

\subsubsection{Keypoints Extraction and Matching}
During scan matching, it is common in practice to use a small subset of the points in the matching process to save computational power. Similar to LOAM \cite{zhang2017low}, we use local surface smoothness to extract keypoints that lie on planar and edge surfaces from the current scan then we match them to planes and lines in the previous scan, respectively. We use KDTree data structure \cite{bentley1975multidimensional} to search in the previous scan for the $K$ nearest neighbors for each keypoint in the current scan. In this work, we fix the number of neighbors $K$ = 5 for points that lie on planes and $K$ = 4 for point that lie on lines. These values were selected using quick manual parameter tuning. Changing the value of $K$ would change the overall framework accuracy and processing time. However, tuning $K$ is not the purpose of this study. Before searching for nearest neighbors of a keypoint $\boldsymbol{p_{curr}}$ in the keypoints of the previous scan, the keypoint $\boldsymbol{p_{curr}}$ is first transformed to the coordinate frame of the LiDAR pose at the previous time step using the latest estimated motion transformation. We search for neighbors in the previous scan only in the subset of keypoints that have the same semantic class (assigned by the deep semantic segmentation model) and surface type (plane or line) as the point in the current scan $\boldsymbol{p_{curr}}$, which increases the probability of finding correct matches and decreases the ratio of outlier matches. To this end, we use a separate KDTree for each semantic class and surface type to make searching for nearest neighbors more efficient. 

\begin{figure}[t]
\begin{subfigure}{0.49\textwidth}
  \includegraphics[width=\linewidth]{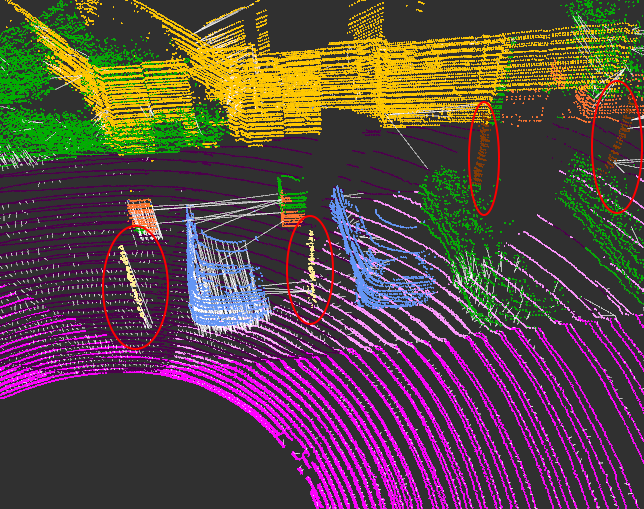}  
  \caption{Without exploiting semantic information}
  \label{fig:nosem_matches}
\end{subfigure}
\begin{subfigure}{0.49\textwidth}
  \includegraphics[width=\linewidth]{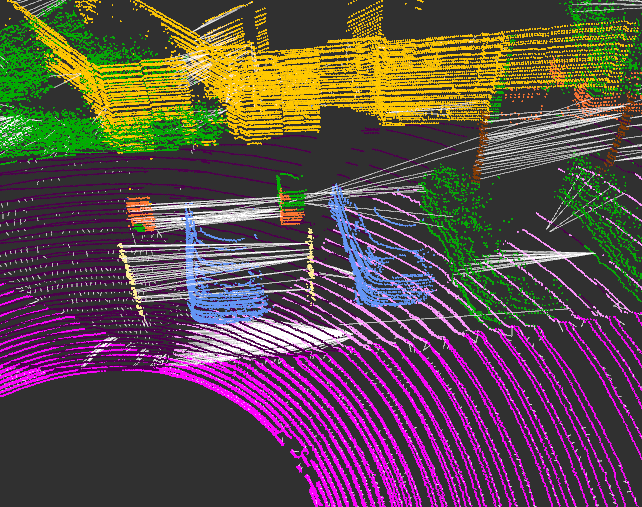}
  \caption{With exploiting semantic information}
  \label{fig:sem_matches}
\end{subfigure}
\caption{Visualizing matches between the first and fifth scans in KITTI sequence 0 with and without exploiting semantic information. Lines are drawn between keypoints in the current scan and the center of their selected neighbors in the previous scan. Points are colored according to their semantic class labels assigned by the deep semantic segmentation model.}
\label{fig:sem_nosem}
\end{figure}

Figure \ref{fig:sem_nosem} shows the effect of using semantic information during matching. In Figure \ref{fig:nosem_matches}, when we do not use semantic information in Algorithm \ref{alg:sem_lidar_odom} during nearest neighbors search, points from the traffic sign pole and the tree truck (annotated by red ellipses in the figure) got matched to the ground instead of being matched to their corresponding objects in the previous scan. This would cause the estimated transformation to have high error with respect to the ground truth motion between the two scans, which will in turn increase the number of outlier matches in the following iteration. Consequently, Algorithm \ref{alg:sem_lidar_odom} will require more iterations before it converges to the true value for the transformation, and it may even fail to converge if the gap between the two scans is too large. Figure \ref{fig:sem_matches} illustrates that by using semantic information, points from the current scan got matched successfully to planes and lines that lie on the same object in the previous scan. Similar to \cite{deschaud2018imls, moreno2020study}, we consider only matches from static objects and discard all points that belong to potentially moving objects to improve the robustness of the framework and enable it to function in dynamic environments. Therefore, points that belong to the car in Figure \ref{fig:sem_matches} were not included in the matching process.

We use Principal Component Analysis (PCA) \cite{jolliffe2016principal} to fit a plane or a line to the local surface implied by the nearest neighbors of each keypoint in the current scan. PCA computes three Eigen vectors, and three corresponding Eigen values that describe the spread of the neighbor points along the direction of each Eigen vector. For planar surfaces, we use the Eigen vector $\boldsymbol{n}$ that corresponds to the smallest Eigen value as a vector perpendicular to the plane (as the two vectors corresponding to the principal axes with the largest Eigen values span the plane). Whereas for edge surfaces, we use the Eigen vector $\boldsymbol{v}$ that corresponds to the largest Eigen value as a director vector that has the same direction as the line. We can then compute the component of any arbitrary vector $\boldsymbol{u}$ along the normal direction to the plane or edge using Equations \ref{eq:residual_planes} and \ref{eq:residual_lines} respectively.
We use the perpendicular distances as residuals in the least squares optimization for motion estimation. In addition, we also use these vectors in our criterion for outlier matches identification as we explain in the next section.

\begin{equation} \label{eq:residual_planes}
\boldsymbol{u_{along\_normal}} = \boldsymbol{n} \boldsymbol{n}^T \boldsymbol{u}
\end{equation}

\begin{equation} \label{eq:residual_lines}
\boldsymbol{u_{along\_normal}} = (I - \boldsymbol{v} \boldsymbol{v}^T) \boldsymbol{u}
\end{equation}

\subsubsection{Outliers Rejection and Motion Estimation}
\begin{figure}[t]
\begin{subfigure}{0.49\textwidth}
  \includegraphics[width=\linewidth]{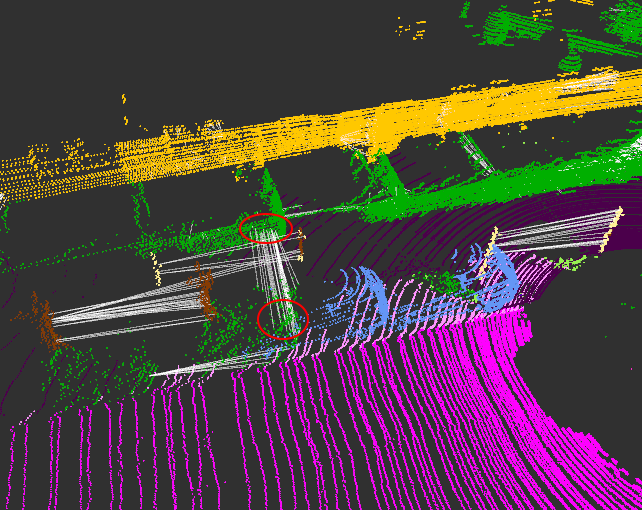}  
  \caption{Matches before outliers rejection}
  \label{fig:no_rejection}
\end{subfigure}
\begin{subfigure}{0.49\textwidth}
  \includegraphics[width=\linewidth]{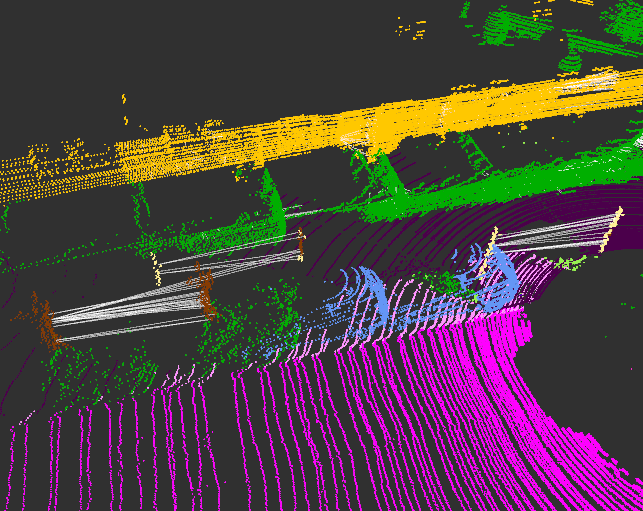}
  \caption{Matches after outliers rejection}
  \label{fig:rejection}
\end{subfigure}
\caption{Visualizing matches between the first and fifth scans in KITTI sequence 0 with and without outliers rejection. Lines are drawn between keypoints in the current scan and the center of their selected neighbors in the previous scan. Points are colored according to their semantic class labels assigned by the deep semantic segmentation model.}
\label{fig:rejection_norejection}
\end{figure}

Exploiting semantic information during matching does not eliminate all outliers from the process. Since two objects of the same semantic type can be near to each other, a point on one object may get matched to a plane or a line on another object of the same class, which is the case with the outlier green vegetation matches in Figure \ref{fig:no_rejection}, where vegetation points from one side got matched with vegetation from the other side of the sidewalk in the other LiDAR scan (annotated by red ellipses in the figure). Such outliers can sometimes be more harmful than the outliers that we get when we do not use semantic information during matching. As we can see in Figure \ref{fig:nosem_matches}, when we do not use semantic information, points from different objects get matched to the ground plane. Nevertheless, if there are enough correct matches, the optimization would still be able to converge to a good estimation of the motion transformation, because points that got matched to the ground do not constrain the estimated horizontal motion. However, when using semantics, false matches between different objects of the same semantic class can pull the estimated transformation during optimization away from the true value.

In this work, we explicitly detect and eliminate outlier matches from the process. The key idea behind our approach is that during optimization each match wants to modify the estimated transformation such that the point moves along the normal direction towards the plane or line to which it was matched. Since in practice the majority of matches are usually correct matches (inliers), points that were falsely matched do not move much towards their corresponding planes and lines. To utilize this fact in outliers detection, we first run the optimization to estimate the motion transformation using all matches in the system. Then for each keypoint $\boldsymbol{p_{curr}}$ in the current scan, we compute the position of $\boldsymbol{p_{curr}}$ relative to the LiDAR pose at time $t - \Delta t$ before and after updating the estimated transformation, where $\boldsymbol{p_{old}} = \mathbb{T}_{initial} * {}^t\boldsymbol{p_{curr}}$, $\boldsymbol{p_{new}} = \mathbb{T} * {}^t\boldsymbol{p_{curr}}$ and $\boldsymbol{u} = \boldsymbol{p_{new}} - \boldsymbol{p_{old}}$. We analyze the vector $\boldsymbol{u}$ along the direction normal to the surface to which it was matched ($\boldsymbol{u_{along\_normal}}$) using Equations \ref{eq:residual_planes} and \ref{eq:residual_lines} for planes and edges respectively, and perpendicular to the direction of the normal ($\boldsymbol{u_{perpendicular\_normal}} = \boldsymbol{u} - \boldsymbol{u_{along\_normal}}$). If the match corresponding to $\boldsymbol{p_{curr}}$ is a correct match, the update in the transformation will cause $\mathbb{T} * {}^t\boldsymbol{p_{curr}}$ to move towards the surface to which it was matched along the direction normal to that surface. 

\sloppy We identify as outliers all points where the ratio $||\boldsymbol{u_{perpendicular\_normal}}|| / ||\boldsymbol{u_{along\_normal}}||$ exceeds some threshold, which corresponds to the absolute tangent of the angle between the direction of $\boldsymbol{u}$ and the direction of the normal to the surface. Points for which matching cost increases after optimization are treated as outliers. However, points corresponding to small costs less than some threshold are always treated as inliers regardless of their motion during transformation. We discard identified outliers then re-solve the optimization with only inlier matches included, while using the latest estimated transformation as an initial guess, then we return to the step of recomputing the nearest neighbor matches for all keypoints after transforming them using the latest estimated transformation. This procedure is repeated until the transformation converges. Algorithm \ref{alg:orme} summarizes our approach for Outliers Rejection and Motion Estimation (ORME). The threshold values $cost_{tol}$ and $r_{tol}$ are hard coded parameters that were tuned using grid search as discussed in Section \ref{hyperparameters_tuning}.

After selecting and refining matches, the function $EstimateTransformation$ in Algorithm \ref{alg:orme} estimates the transformation that minimizes the mean of squared perpendicular distances over all point-to-line and point-to-plane matches. Similar to \cite{zhang2017low}, we use Levenberg-Marquardt algorithm to solve the least squares optimization problem. We refer the reader to \cite{zhang2017low} for more details about odometry estimation. 

Figure \ref{fig:rejection} shows that our outliers rejection algorithm was able to detect and eliminate false matches corresponding to green vegetation that do not agree with the direction of motion implied by the majority of matches in the system. Although our outliers rejection process can add extra computational overhead to the system, we show in Section \ref{sec:processing_time} that it can in some cases improve the overall processing time, because the inclusion of outlier matches in the optimization can require more optimization iterations before the transformation converges. We also show in Section \ref{sec:quantitative} that outliers rejection can improve the framework robustness against high speed motion.

\begin{algorithm}[tb]
\footnotesize
\begin{algorithmic}[]
\State \textbf{Initialize} $\mathbb{T} \gets \mathbb{T}_{initial}$
\State \textbf{Initialize} $iter \gets 1$
\texttt{\\}

\While{no convergence of $\mathbb{T}$ and $iter \leq iters_{max}$}

    \State // Initialize the set of inlier matches as an empty set
    \State $\mathbb{M} \gets \emptyset$\\

    \For {match $({}^t\boldsymbol{p_{curr}}, planeOrLine) \in \mathbb{M}_{initial}$}
        \State // For the first iteration, add all matches as inliers
        \If{$iter == 1$} 
            \State $\mathbb{M} \gets \mathbb{M}  \cup  \{({}^t\boldsymbol{p_{curr}}, planeOrLine)\}$
            \State $\textbf{continue}$
        \EndIf
        \texttt{\\}
        
        \State // Compute the position of $\boldsymbol{p_{curr}}$ relative to the LiDAR pose at time $t - \Delta t$ 
        \State // before and after updating the estimated $\mathbb{T}$
        
        \State $\boldsymbol{p_{old}} \gets \mathbb{T}_{initial} * {}^t\boldsymbol{p_{curr}}$ 
        \State $\boldsymbol{p_{new}} \gets \mathbb{T} * {}^t\boldsymbol{p_{curr}}$ 
        \State $\boldsymbol{u} \gets \boldsymbol{p_{new}} - \boldsymbol{p_{old}}$ 
        \texttt{\\}
        
        \State // Analyze the vector $\boldsymbol{u}$ 
        \State $ u_{along\_normal} \gets $ Component of $\boldsymbol{u}$ along the normal to the plane or line
        \State $ u_{perpendicular\_normal} \gets $ Component of $\boldsymbol{u}$ perpendicular to the normal to the plane or line 
        \State $ratio \gets u_{perpendicular\_normal}/u_{along\_normal}$
        \texttt{\\}
        
        \State // Compute costs using Equations \ref{eq:residual_planes} and \ref{eq:residual_lines} 
        \State // The cost of a match is the squared magnitude of its corresponding residual
        \State $cost_{old} \gets $ perpendicular distance squared from $\boldsymbol{p_{old}}$ to the plane or line
        \State $cost_{new} \gets $ perpendicular distance squared from $\boldsymbol{p_{new}}$ to the plane or line
        \texttt{\\}
        
        \State // All points with $cost < cost_{tol}$ are treated as inliers 
        \State // Otherwise, add only matches that satisfy the criteria 
        \If{$cost_{new} < cost_{tol}$ or ($ratio < r_{tol}$ and $cost_{new} < cost_{old}$)} 
            \State $\mathbb{M} \gets \mathbb{M}  \cup  \{({}^t\boldsymbol{p_{curr}}, planeOrLine)\}$
        \EndIf
    \EndFor
    \texttt{\\}

    \State $\mathbb{T} \gets EstimateTransformation(\mathbb{T}, \mathbb{M})$ 
    \State $iter$ += 1 
\EndWhile

\texttt{\\}
\State return $\mathbb{T}, \mathbb{M}$
\end{algorithmic}

\caption{$ORME(\mathbb{T}_{initial}, \mathbb{M})$}
\label{alg:orme}
\end{algorithm}

\subsection{Semantic Localization and Mapping}
After estimating the relative motion between the two scans, the estimated motion transformation is applied to the pose estimation of the previous time step to yield the initial pose estimation for the current time step. This pose estimation is further refined by registering the current scan to the point cloud aggregated over the past time steps, which serves as a global map of the environment. Global poses are relative to the coordinate frame of the LiDAR at the time of the first scan.

\subsubsection{Scan to Map Registration}
Similar to what we did in the scan-to-scan registration for motion estimation, we exploit the semantic information in the scan-to-map localization. To this end, we extract the local region of the map surrounding the estimated current pose estimation, then split the local map into separate point clouds for each semantic class and surface type. We then construct a separate KDTree for each (semantic class, surface type) pair and try to match each keypoint in the current scan to a surface in the map in a similar way to what we did in Section \ref{sec:semantic_odometry}. We again solve the least-squares problem to find the pose transformation that minimizes the cost over the matches between keypoints in the current scan and their corresponding planes and lines in the map. The output pose represents the final pose estimation for the current time step.

\subsubsection{Map Update}
The map is initialized as an empty point cloud, then updated at each time step by adding to it points from the registered LiDAR scan at that time step. The final estimated pose for the current time step is used to transform all points in the current scan from the coordinate frame defined by the current LiDAR pose to the world coordinate frame, which is defined by the LiDAR pose at the very first time step. 

To build a complete semantic LiDAR SLAM, we need to merge points from different semantic classes into a single map. To best of our knowledge, all existing semantic LiDAR SLAM approaches store a separate map for each semantic class, and perform simple concatenation of these maps to get a single semantic map of the environment. The problem with this approach is that as the platform moves, it may observe the same object more than once. As the deep learning semantic segmentation model can generate inaccurate segmentations for some points, we may find that the same object is represented more than once in the map with different semantic classes, which would lead to more outlier matches during scan-to-map registration.

To this end, transformed points from all classes from the current scan are concatenated into a single point cloud then appended to the list of points representing the map point cloud. We propose to use a custom voxel-grid downsampling filter that discretizes the 3D space into voxels of predefined side length. For each voxel, we compute the centroid of all points lying within the voxel. The output of downsampling is the point cloud containing the centroids from all voxels. We assign a semantic label for each centroid point of each voxel, which is the label of the point in the voxel that had the minimum euclidean distance to the LiDAR when the point was observed. The reason for this choice is that points captured near the LiDAR sensor have more precise location, and less noisy semantic labels from the deep semantic segmentation network as reported by our previous work \cite{DBLP:journals/corr/abs-2011-01974}. 

\section{Experiments and Results} \label{sec:results}
In this section, we present the experimental evaluation of our proposed framework. The goal of these experiments is to show the value of improving keypoints matching by exploiting semantic information and our novel outliers rejection algorithm in improving the quality of trajectory estimation. We also visualize samples of the semantic maps constructed using our framework.

\subsection{Dataset}
\label{sec:dataset}
We evaluate our framework on KITTI odometry dataset \cite{geiger2012we}. The dataset consists of over 43000 Velodyne HDL-64E LiDAR scans, divided  into  11  training  sequences  for  which  ground-truth poses (computed by fusion of GPS and IMU) are provided, and 11 test sequences. We trained both spherical view and bird's-eye view networks on SemanticKITTI dataset \cite{behley2019semantickitti} which provided point-wise semantic label annotations for all scans in the KITTI odometry dataset \cite{geiger2012we}. We train our deep semantic segmentation network on the even training sequences 00, 02, 04, 06, 10. And validate it on sequence 08. We used the same hyper-parameter setting as in \cite{DBLP:journals/corr/abs-2011-01974}. The odd sequences 01, 03, 05, 07, 09 are used to evaluate our proposed LiDAR odometry and mapping framework and compare it against other approaches.

To test our localization accuracy, we use the evaluation devkit \footnote{\label{footnote1}\url{http://cvlibs.net/datasets/kitti/eval_odometry.php}} provided by KITTI benchmark to compute the average error. They extract overlapping trajectories of lengths 100m, 200m, ..., 800m from each sequence. For each trajectory, the error is computed as the translation error between the translation computed from the ground truth trajectory and the translation computed from the estimated trajectory. The error for each trajectory is expressed as a percentage of the trajectory length. The overall error for a sequence is simply the average over these trajectory percentage errors. More information can be found at \cite{geiger2012we}.

Although the sequences in KITTI dataset exhibit various car speeds, we want to evaluate our framework in more extreme driving speeds. However, since we have no control over the driving speed in KITTI dataset, we simulate different vehicle speeds by dropping intermediate LiDAR scans between every two processed scans. The intuition behind why dropping scans resembles high speed motion is that when the moving platform has high speed relative to the scanning frequency of the LiDAR, the distance between the LiDAR position at two consecutive LiDAR scans is large. Similarly, dropping intermediate scans from KITTI dataset simulates high speed motion by performing registration between LiDAR scans that are far from each other in time and space. In this work, we compare different approaches while dropping 0, 1, ..., 10 intermediate scans between every two processed scans.

\subsection{Hyper-parameters Tuning} \label{hyperparameters_tuning}
\begin{figure}[tb]
    \centering
	\includegraphics[height=10cm, keepaspectratio]{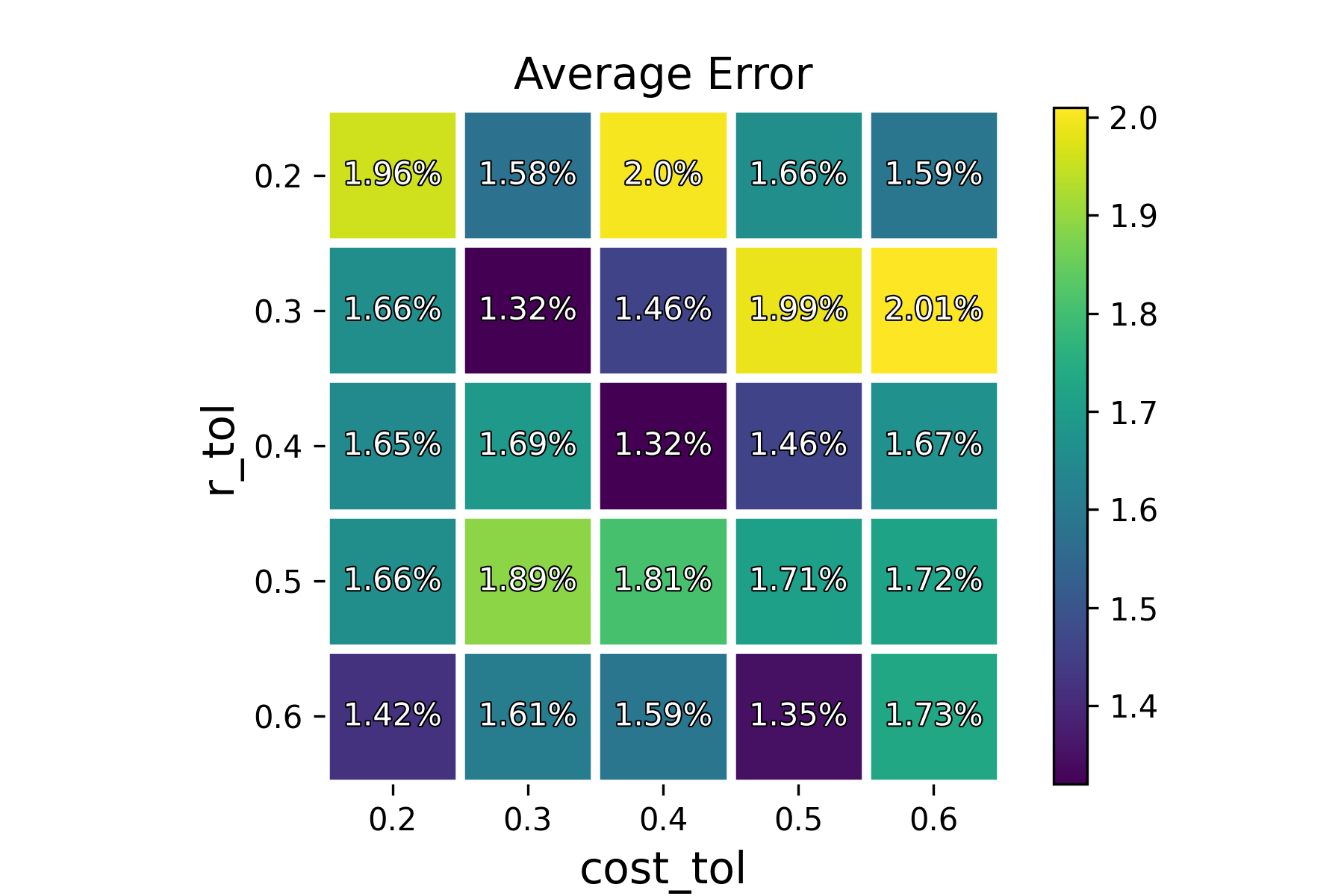}
	\vspace{3pt}
	\caption{Heat map for tuning the parameters of the outliers rejection algorithm}
	\label{fig:heat_map}
\end{figure}

We use grid search to select the best values for the outliers rejection algorithm parameters ($r_{tol}$ and $cost_{tol}$). Figure \ref{fig:heat_map} shows the average error over KITTI odd sequences for different values of the two parameters while skipping 10 intermediate scans between each two processed scans. Setting $r_{tol}$ = 0.4 and $cost_{tol}$ = 0.4 achieves the best average error of $1.32\%$. We fix these values in the subsequent experiments, which shows that these values also generalize well on different numbers of skipped scans.

LOAM performance is sensitive to the choice and tuning of the system hyper-parameters such as the downsampling filter sizes, the parameters of the loss function that assigns weights to residuals to reduce the effect of outliers, and the parameters of the keypoints extraction module. Unfortunately, the official implementation of LOAM is not publicly available, and the values of the parameters used by the authors were not provided in their paper \cite{zhang2017low}. Tuning these parameters to achieve the best performance on KITTI dataset is not the purpose of this work, and we rather focus on studying the effect of exploiting semantic information and outliers rejection on localization.




\subsection{Experimental Evaluation}
In our experiments, we evaluate the performance of bare LOAM architecture without exploiting semantic information, LOAM + semantic matching, and LOAM + semantic matching + outliers rejection. During all experiments, we use M-estimation techniques for robust motion estimation. We use Huber loss for 4 iterations and Arctan loss for another 4 iterations during odometry estimation, and use only 4 iterations of Arctan loss during scan-to-map localization. 

We compare our framework against SUMA++ \cite{chen2019suma++}, which is the only open source framework on KITTI benchmark \footnote{Same link as in footnote \ref{footnote1}} that exploits semantic information in LiDAR odometry. We use the official SUMA++ implementation on github \footnote{ \url{https://github.com/PRBonn/semantic_suma/tree/ca2d275729bbaa9b87b784793c191a8714a46d16}}. We increase the maximum number of ICP iterations in SUMA++ to 100 iterations to allow the optimization to converge as we drop intermediate scans (because the registered scans become far from each other when we drop scans). When we drop 5 and 10 intermediate scans, we increase the value of the parameter ``icp-max-distance'' in SUMA++ from 2.0 to 12 and 22 respectively to allow far matches as the gap between consecutive scans gets bigger. When the estimated motion exceeds a certain limit, SUMA++ falls back to performing frame-to-frame instead of frame-to-model matching. We increase the limit from 0.4 meters translation and 0.1 radians rotation to 2.4 meters and 0.6 radians when we drop 5 scans, and 4.4 meters and 1.1 radians when we drop 10 scans (we multiply all parameters by the number of skipped scans + 1). Similarly, we set the maximum distance between a point and its match in our framework to (3 * number of skipped scans + 1) and reject all matches exceeding this distance. These parameters values were selected using quick manual parameter tuning. Most SLAM systems have such parameters that are tuned according to the type of the environment and the expected vehicle speed. The goal of this work is to study the effect of improving the matching process. However, making the system robust to different speeds and operation conditions without having to manually adjust the system parameters can be a very important direction for future work as we discuss in Section \ref{sec:future}.

\subsubsection{Qualitative Evaluation}
\label{sec:qualitative}

\begin{figure}[!htp]
\begin{subfigure}{\textwidth}
  \centering
  \includegraphics[width=0.9\textwidth, height=7cm]{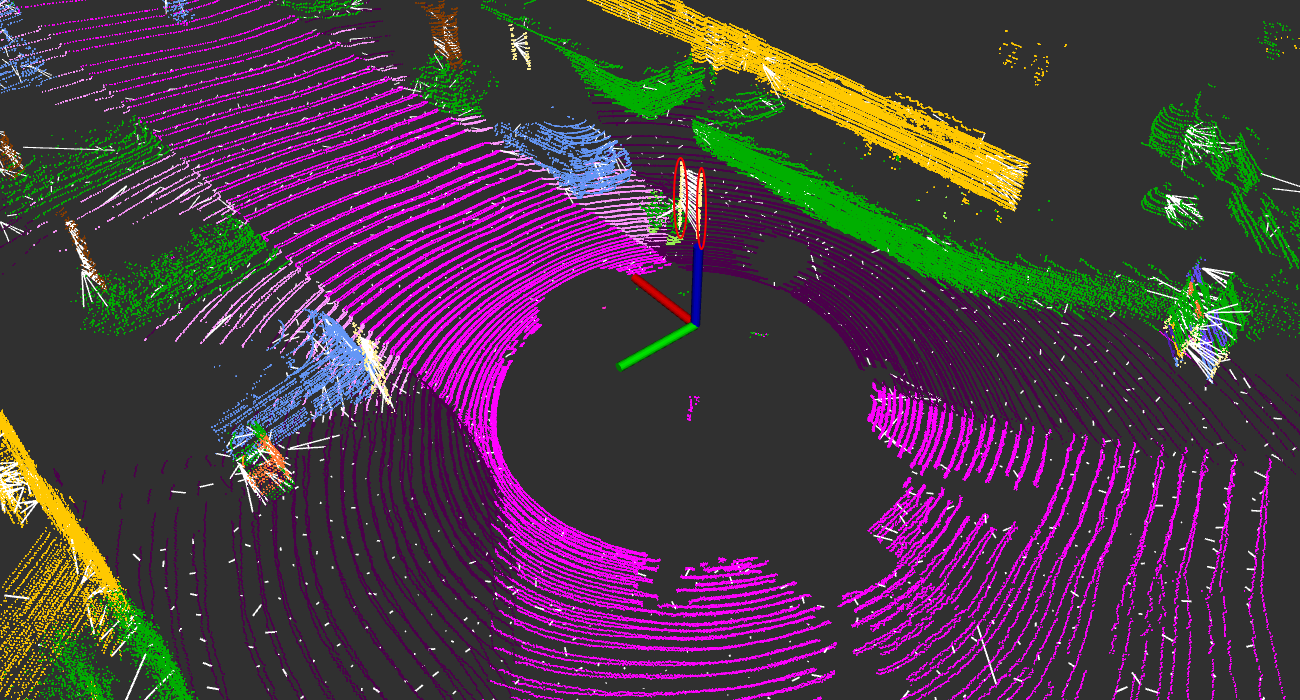}  
  \caption{Initial matches before motion estimation}
  \label{fig:nosem1_a}
  \vspace{10pt}
\end{subfigure}

\begin{subfigure}{\textwidth}
  \centering
  \includegraphics[width=0.9\textwidth, height=7cm]{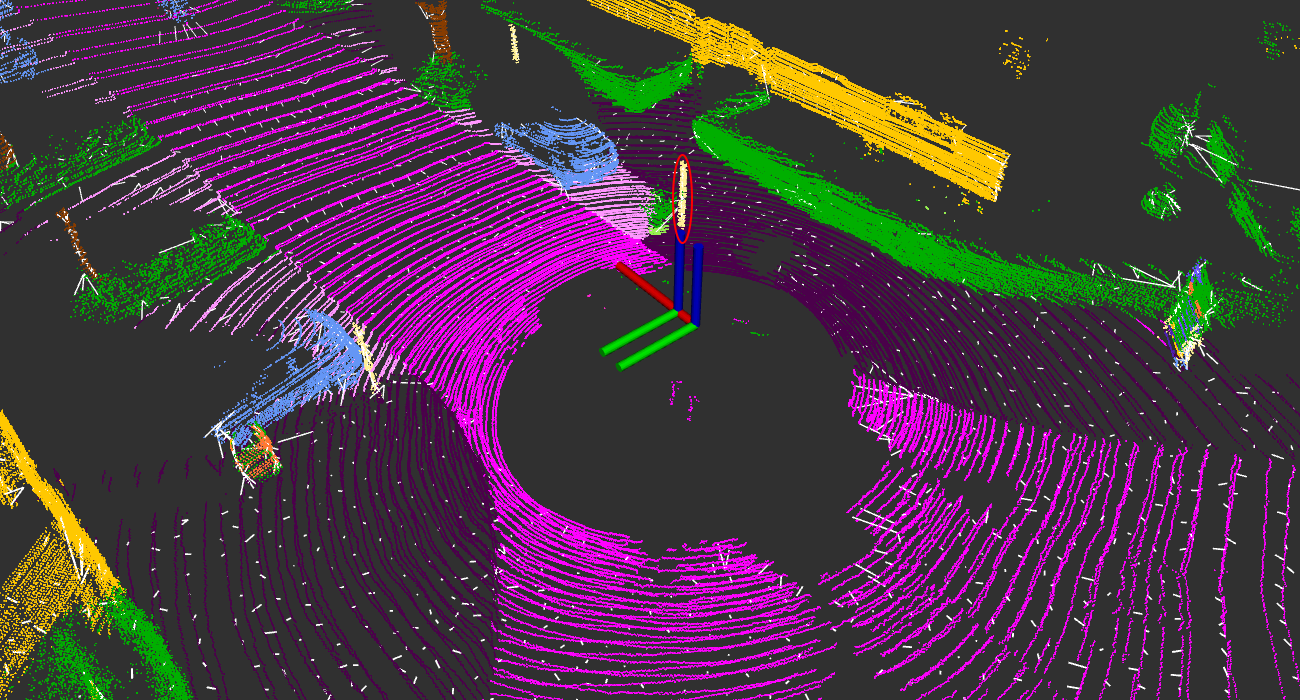}  
  \caption{After motion estimation using the initial matches}
  \label{fig:nosem1_b}
  \vspace{10pt}
\end{subfigure}

\begin{subfigure}{\textwidth}
  \centering
  \includegraphics[width=0.9\textwidth, height=7cm]{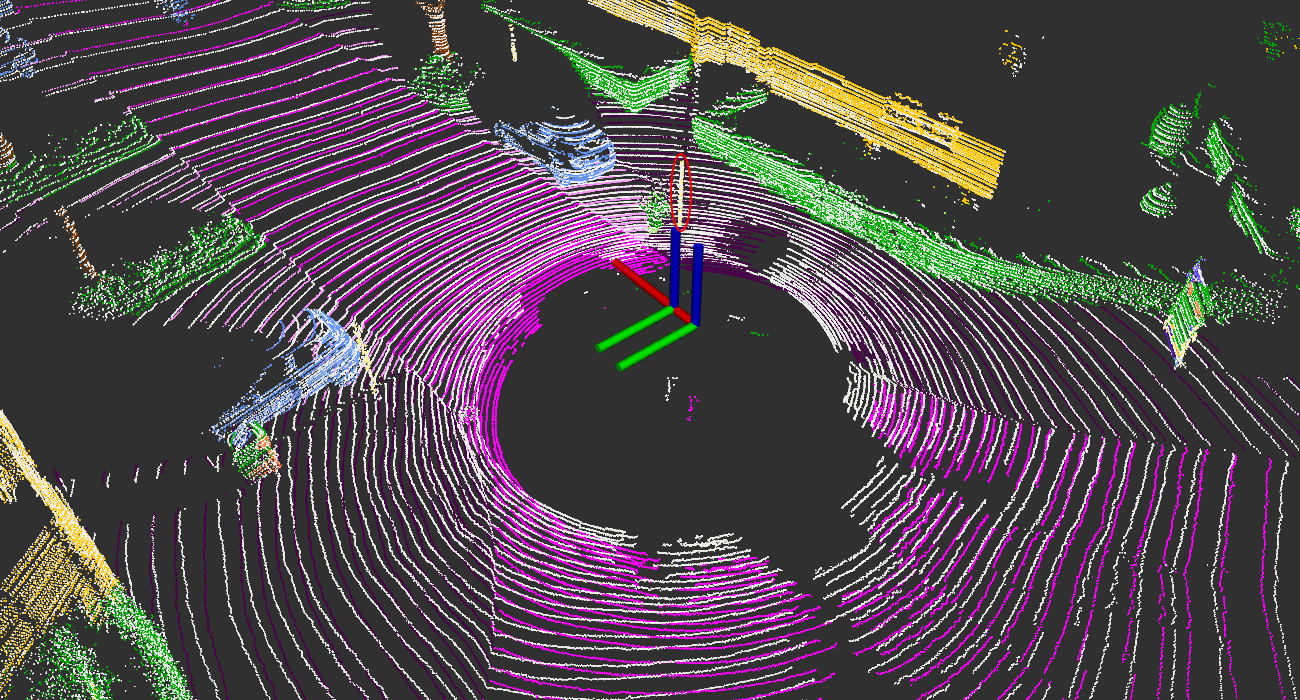}  
  \caption{Final registration}
  \label{fig:nosem1_c}
\end{subfigure}

\caption{Motion estimation between scan $0$ and scan $1$ in KITTI sequence $00$ without using semantics or outliers rejection.}
\label{fig:nosem1}
\end{figure}

We start by visualizing the impact of improving matches using semantic information and outliers rejection on the quality of scan registration and motion estimation. In Figure \ref{fig:nosem1}, we perform scan matching between scan $0$ and scan $1$ in sequence $00$ in KITTI dataset. Figure \ref{fig:nosem1_a} shows the initial point-to-line and point-to-plane matches using nearest neighbors search irrespective of the semantic labels of the points and without rejecting any matches using our outliers rejection algorithm. Matches are visualized as white lines drawn between keypoints in scan $0$ and the center of their selected neighbors in scan $1$. Points are colored in the figure for better appearance according to their semantic labels although these labels are not used during matching. We annotate a traffic sign pole (yellow points) by a red ellipse in the two scans to show the quality of registration. The closer the two ellipses, the better the registration is. Figure \ref{fig:nosem1_b} visualizes the motion estimated by point-to-line and point-to-plane registration using the set of matches selected without exploiting semantic information as visualized in Figure \ref{fig:nosem1_a}. The algorithm is given sufficient number of iterations to fully converge. Figure \ref{fig:nosem1_c} visualizes the final pose after $8$ scan-to-scan motion estimation passes from Algorithm \ref{alg:sem_lidar_odom} followed by $4$ iterations of scan-to-map pose refinement (the map is equivalent to scan $0$ for this demonstration because we process two scans only). We re-color scan $1$ by assigning a white color to all its points to allow the reader visually inspect the quality of registration. We also visualize the initial and final poses using coordinate frame axes. We can see that the two scans have high initial overlap (due to slow vehicle motion / no frame skipping). And therefore even without using semantic information, matches are good enough to estimate the motion transformation.

\begin{figure}[!htp]

\begin{subfigure}{\textwidth}
  \centering
  \includegraphics[width=0.9\textwidth, height=7cm]{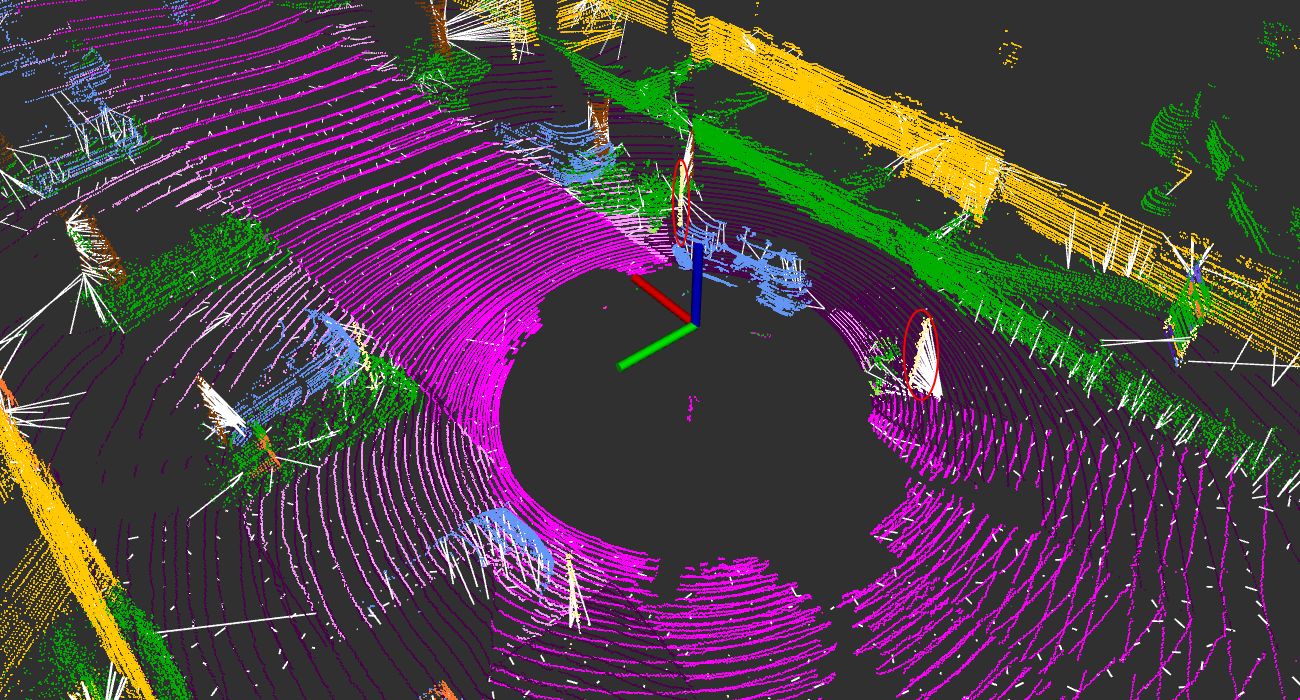}  
  \caption{Initial matches before motion estimation}
  \label{fig:nosem2_a}
  \vspace{10pt}
\end{subfigure}

\begin{subfigure}{\textwidth}
  \centering
  \includegraphics[width=0.9\textwidth, height=7cm]{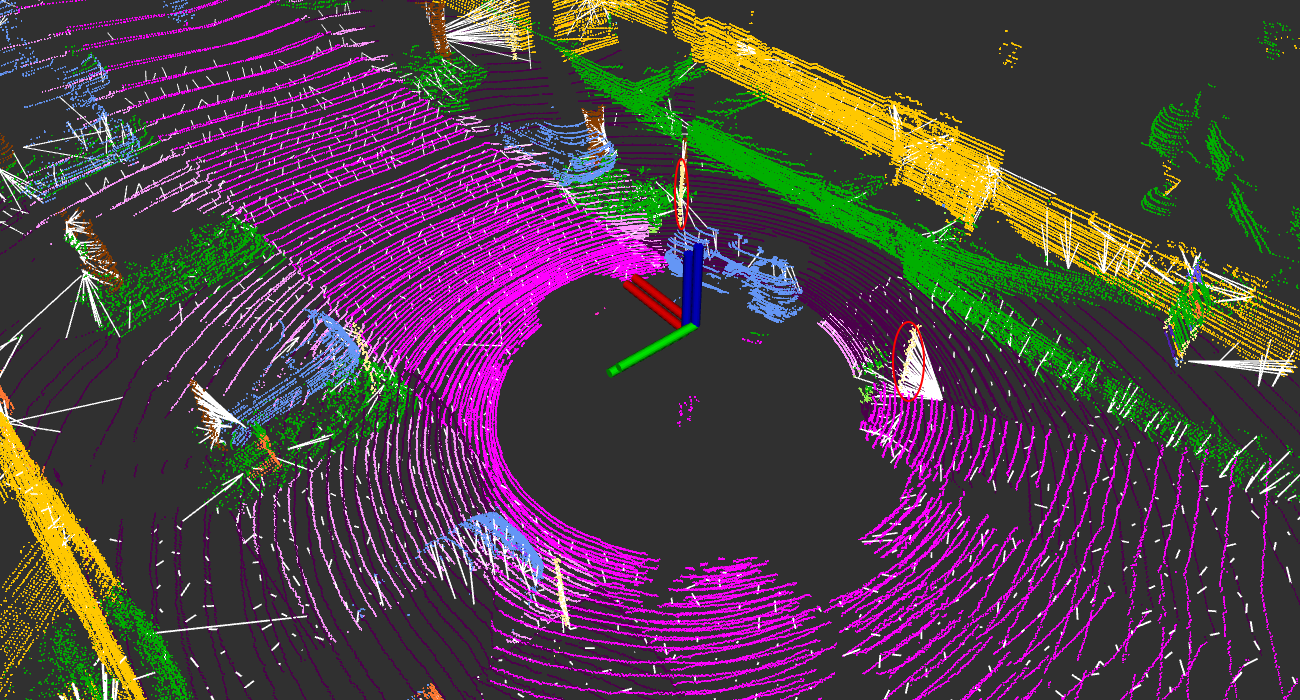}  
  \caption{After motion estimation using the initial matches}
  \label{fig:nosem2_b}
  \vspace{10pt}
\end{subfigure}

\begin{subfigure}{\textwidth}
  \centering
  \includegraphics[width=0.9\textwidth, height=7cm]{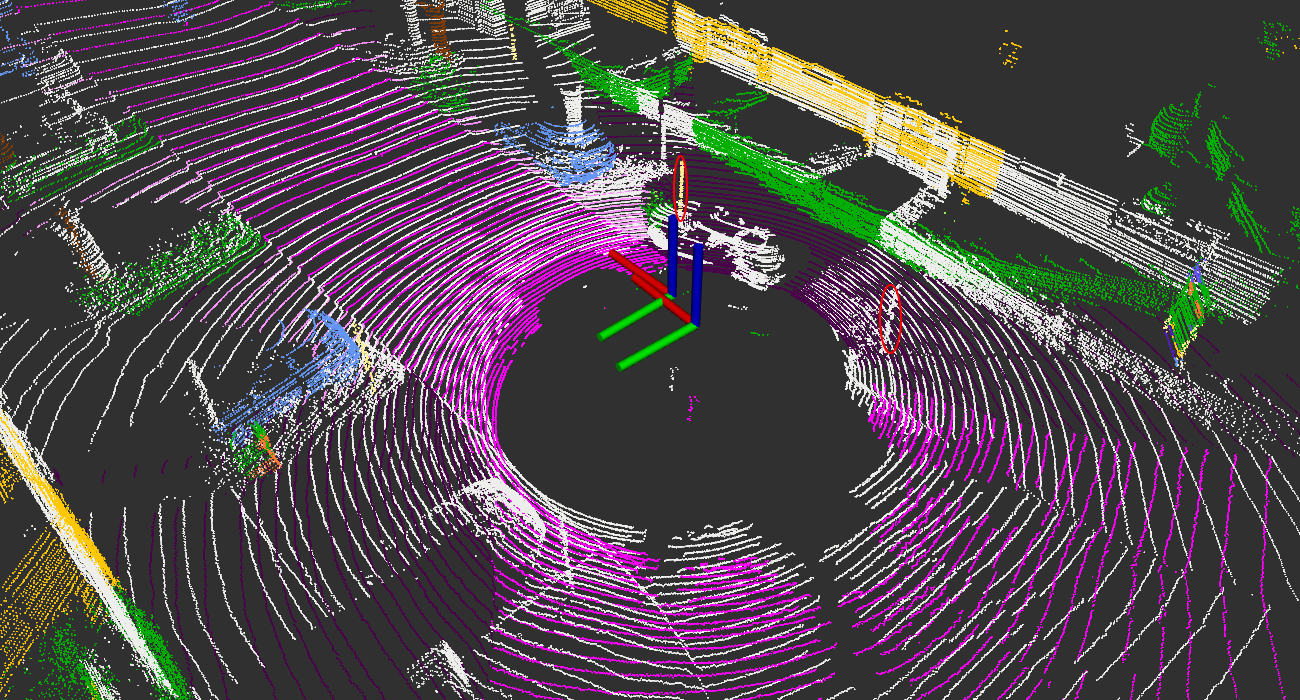}  
  \caption{Final registration}
  \label{fig:nosem2_c}
\end{subfigure}

\caption{Motion estimation between scan $0$ and scan $10$ in KITTI sequence $00$ without using semantics or outliers rejection.}
\label{fig:nosem2}
\end{figure}

\begin{figure}[!htp]

\begin{subfigure}{\textwidth}
  \centering
  \includegraphics[width=0.9\textwidth, height=7cm]{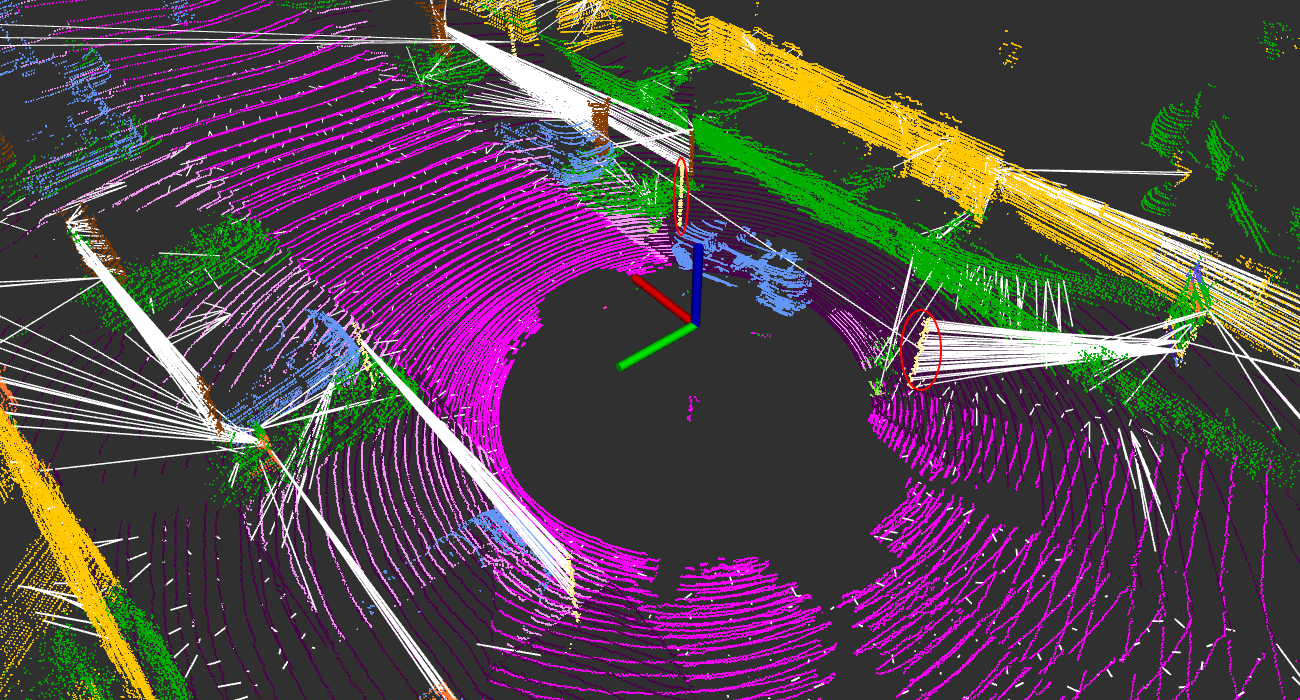}  
  \caption{Initial matches before motion estimation}
  \label{fig:sem_a}
  \vspace{10pt}
\end{subfigure}

\begin{subfigure}{\textwidth}
  \centering
  \includegraphics[width=0.9\textwidth, height=7cm]{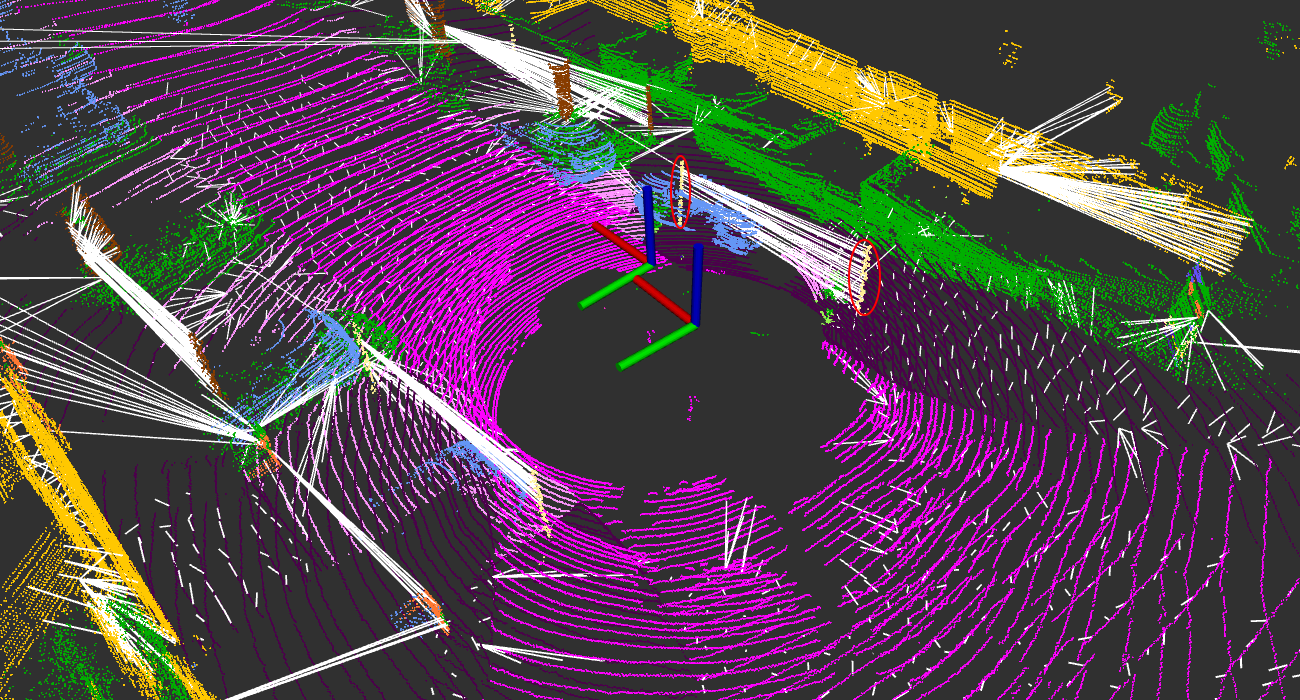}  
  \caption{After motion estimation using the initial matches}
  \label{fig:sem_b}
  \vspace{10pt}
\end{subfigure}

\begin{subfigure}{\textwidth}
  \centering
  \includegraphics[width=0.9\textwidth, height=7cm]{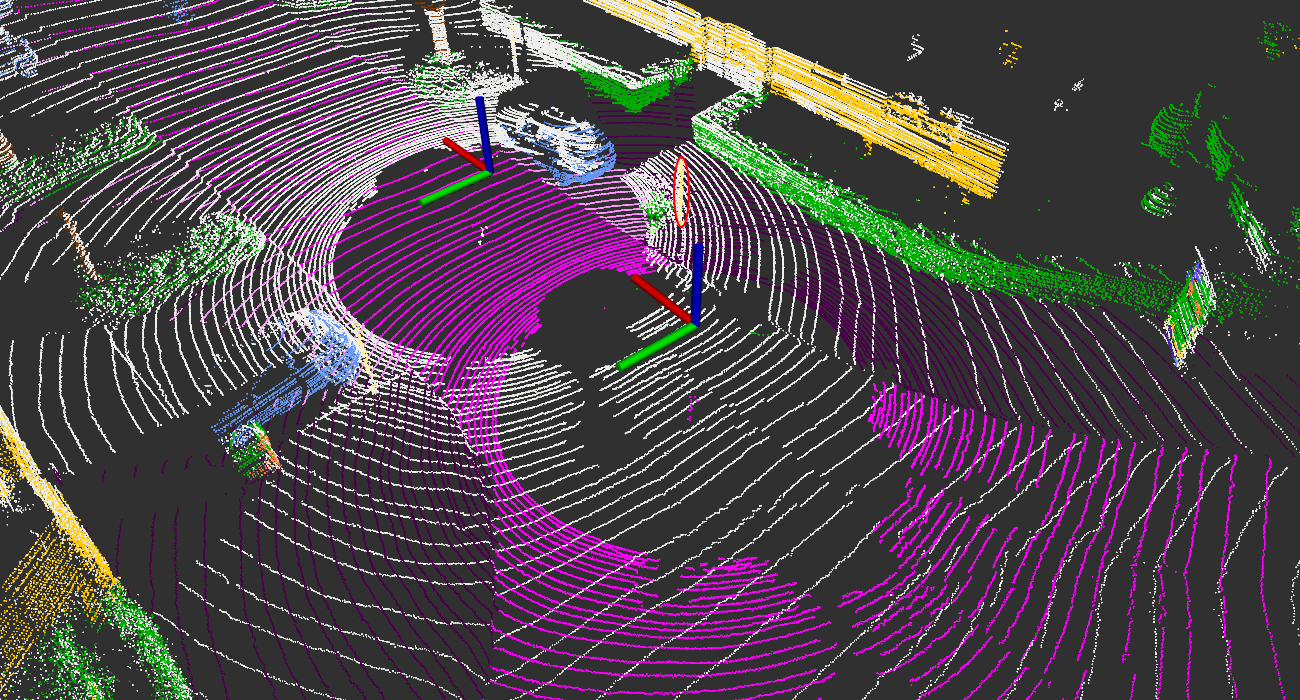}  
  \caption{Final registration}
  \label{fig:sem_c}
\end{subfigure}

\caption{Motion estimation between scan $0$ and scan $10$ in KITTI sequence $00$. Semantic information is exploited during keypoints matching. No outliers rejection is used.}
\label{fig:sem}
\end{figure}

As we drop intermediate scans between the two processed scan to simulate high speed motion, selected matches using nearest neighbors search are no longer good enough to estimate the motion transformation that registers the two point clouds correctly. This is shown in Figure \ref{fig:nosem2} where the algorithm fails to estimate the motion between scan $0$ and scan $10$ in KITTI sequence $00$. When we take semantic information into account, we get better matches as visualized in Figure \ref{fig:sem_a}. The motion estimation estimated in Figure \ref{fig:sem_b} still has high error. However, this error decreases as we iteratively re-compute matches using semantic nearest neighbors search and re-estimate the motion (Algorithm \ref{alg:sem_lidar_odom}) until the transformation converges as shown in Figure \ref{fig:sem_c}. We still use $8$ scan-to-scan passes and $4$ scan-to-map passes. As opposed to Figure \ref{fig:nosem2_c}, the transformation converged to a good value while consuming less time as we report in Section \ref{sec:processing_time}.

\begin{figure}[!htp]

\begin{subfigure}{\textwidth}
  \centering
  \includegraphics[width=0.9\textwidth, height=7cm]{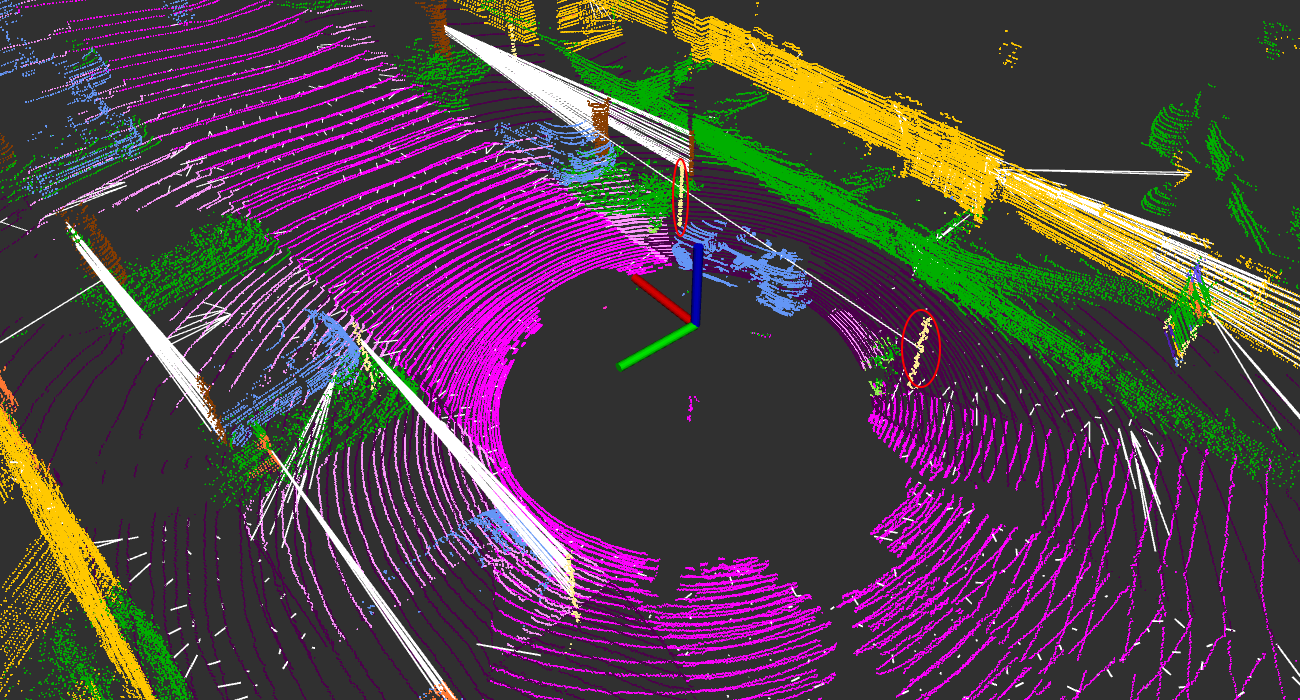}  
  \caption{Initial matches before motion estimation}
  \label{fig:rej_a}
  \vspace{10pt}
\end{subfigure}

\begin{subfigure}{\textwidth}
  \centering
  \includegraphics[width=0.9\textwidth, height=7cm]{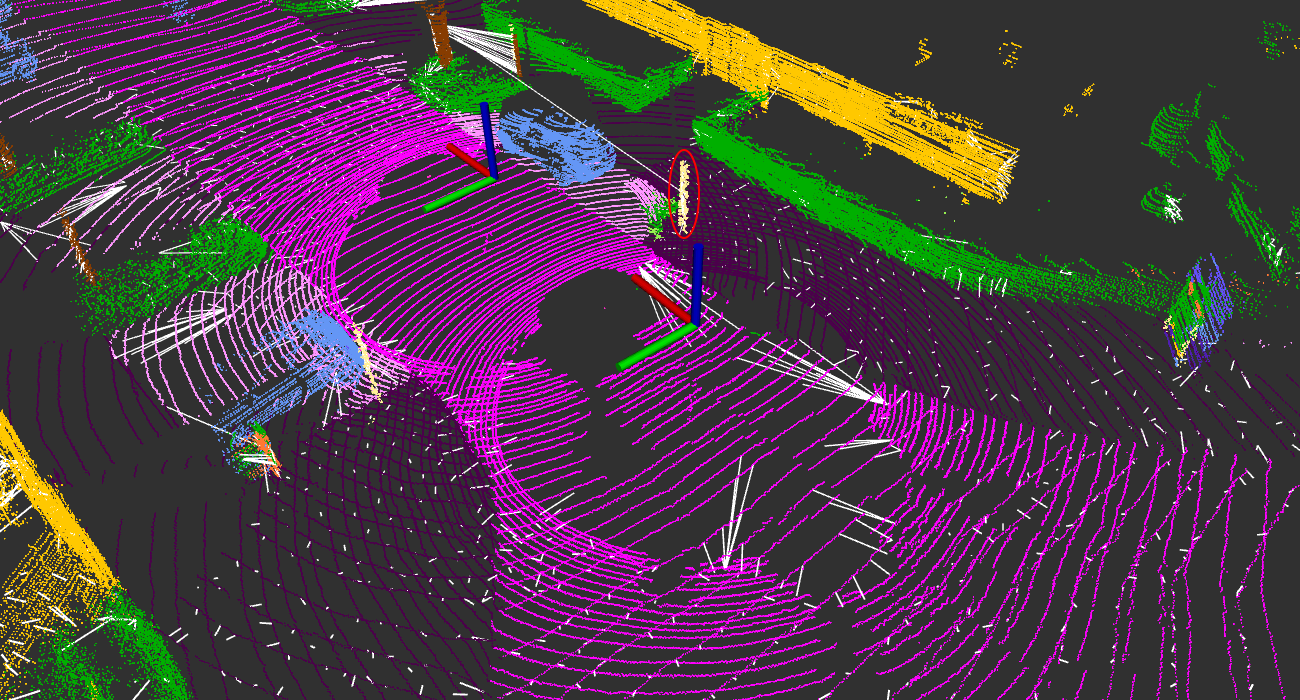}  
  \caption{After motion estimation using the initial matches}
  \label{fig:rej_b}
  \vspace{10pt}
\end{subfigure}

\begin{subfigure}{\textwidth}
  \centering
  \includegraphics[width=0.9\textwidth, height=7cm]{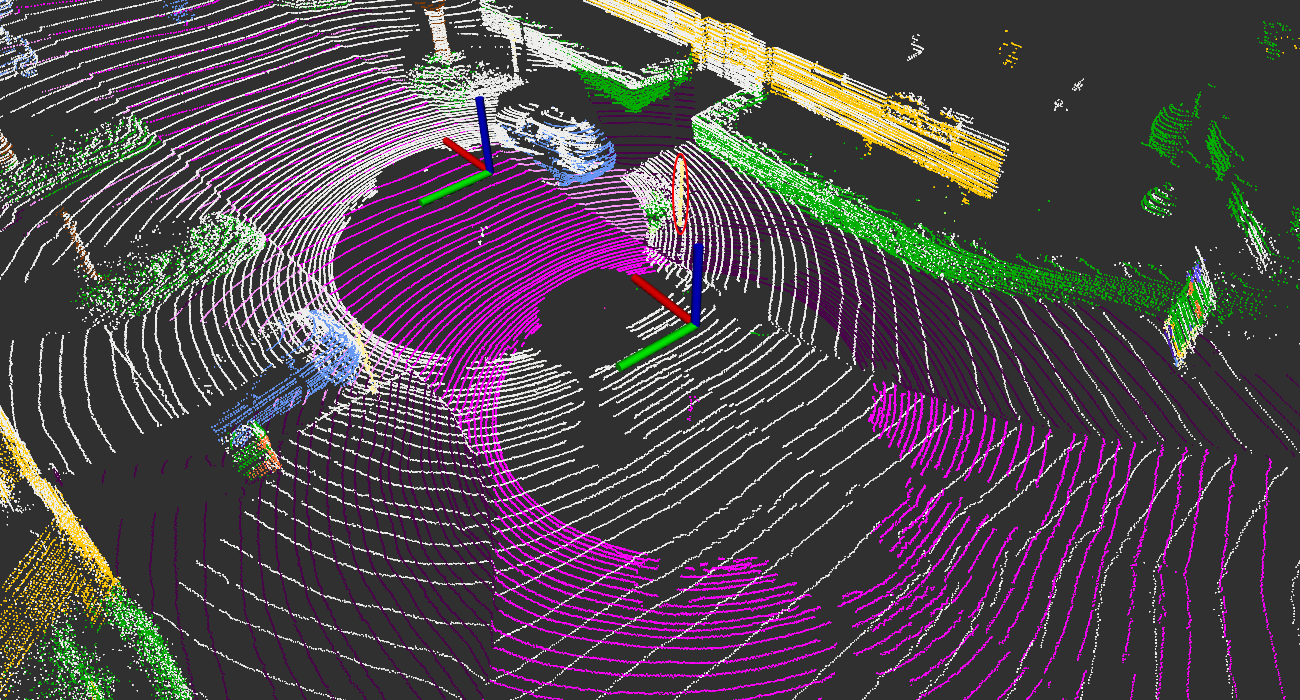}  
  \caption{Final registration}
  \label{fig:rej_c}
\end{subfigure}

\caption{Motion estimation between scan $0$ and scan $10$ in KITTI sequence $00$. Semantic information is exploited during keypoints matching. Matches are further refined using our outliers rejection algorithm}
\label{fig:rej}
\end{figure}

We can see that outlier matches pulled the estimated transformation a little bit to the right in Figure \ref{fig:sem_b}, and even hindered the optimizer from converging to a good transformation value from the first ICP pass. On the other hand, when we use our novel outliers rejection technique, matches get further refined as shown in Figure \ref{fig:rej_a} (compared to figure \ref{fig:sem_a}). The estimated motion from a single ICP pass (without re-computing matches) is shown to be significantly better in Figure \ref{fig:rej_b} than in Figure \ref{fig:sem_b}. This however comes at an extra computational cost, as our outliers rejection algorithm employs internal iterations. We discuss the impact of outliers rejection on trajectory estimation accuracy and processing time in the following sections.

\begin{figure*}[!htbp]
\centering
\begin{subfigure}{\textwidth}
    \centering
    \includegraphics[width=\textwidth]{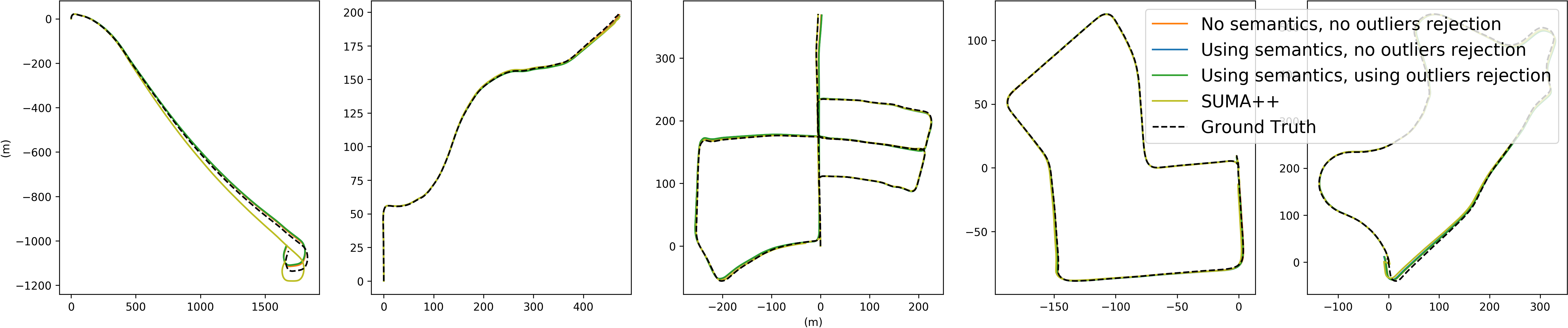}
    \caption{Number of skipped scans = 0}
    \vspace{1cm}
\end{subfigure}

\centering
\begin{subfigure}{\textwidth}
    \centering
    \includegraphics[width=\textwidth]{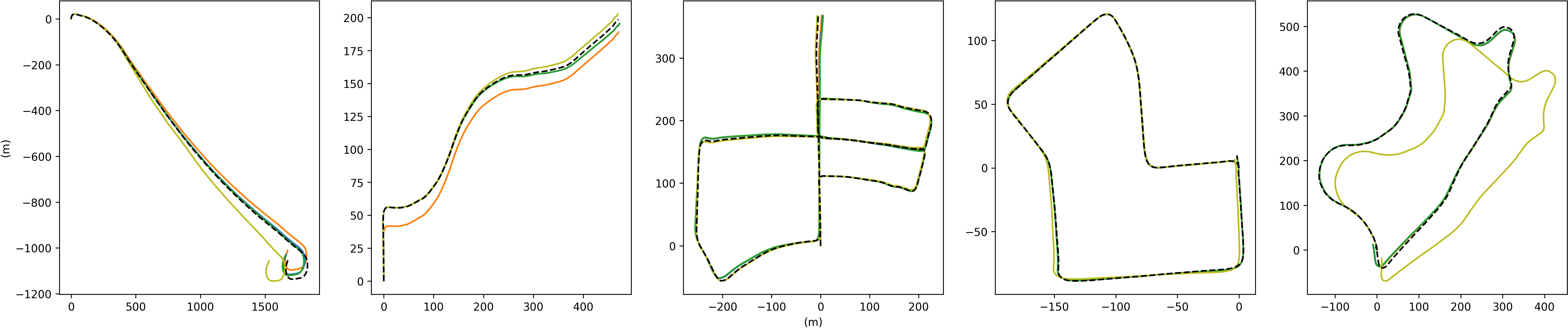}
    \caption{Number of skipped scans = 5}
    \vspace{1cm}
\end{subfigure}

\centering
\begin{subfigure}{\textwidth}
    \centering
    \includegraphics[width=\textwidth]{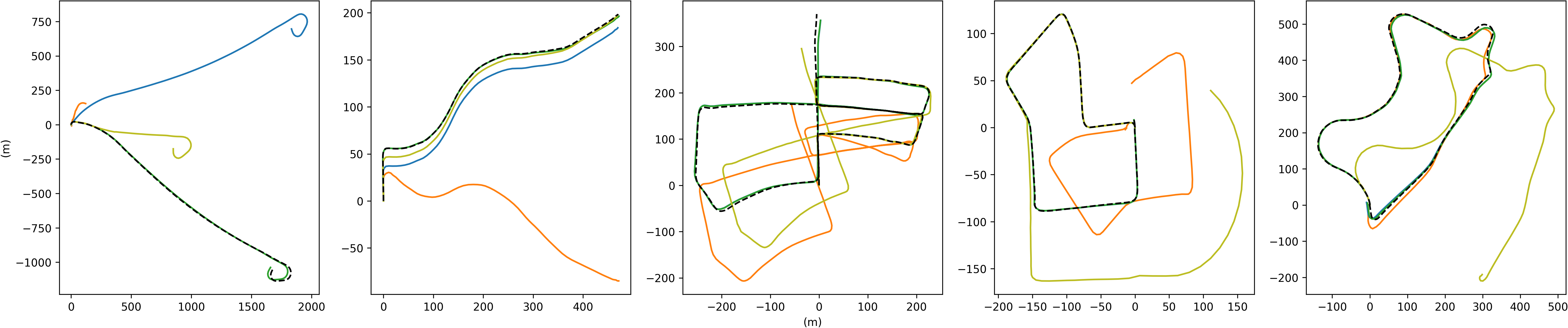}
    \caption{Number of skipped scans = 10}
\end{subfigure}

\caption{Trajectories of different methods on odd KITTI sequences}
\label{fig:trajectories}
\end{figure*}

\begin{figure}[!htbp]
\begin{subfigure}{\textwidth}
    \centering
    \includegraphics[width=\textwidth, height=10cm]{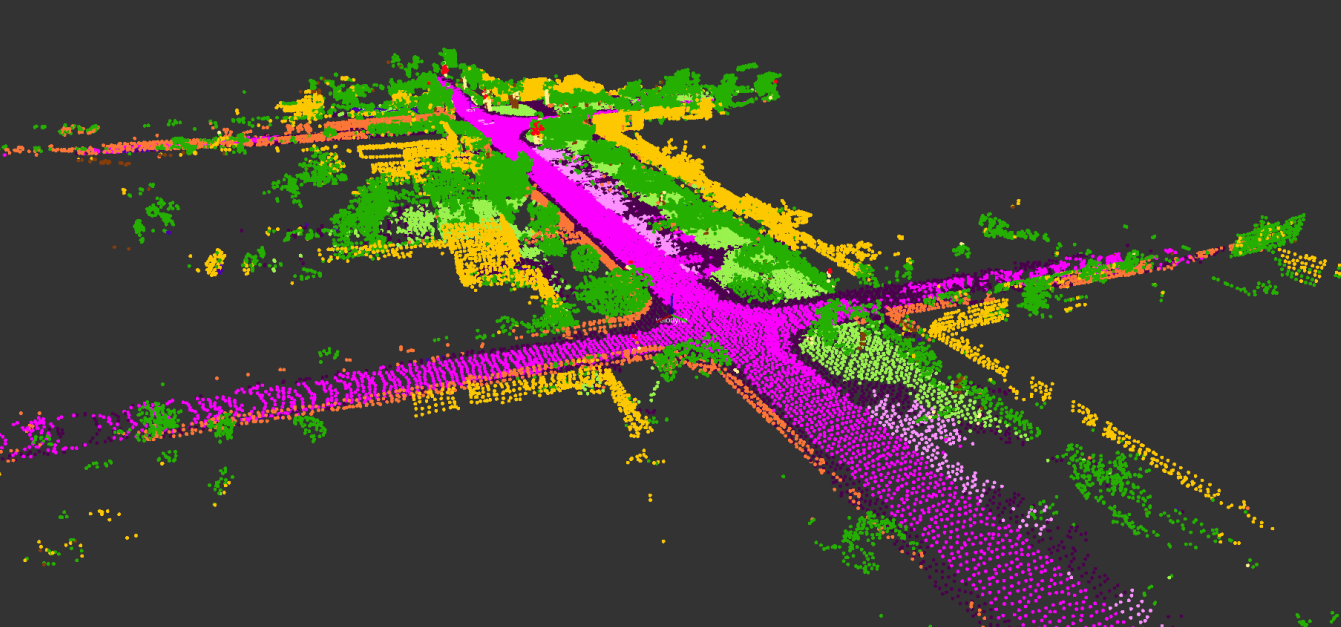}
    \vspace{3pt}
\end{subfigure}

\begin{subfigure}{\textwidth}
    \centering
    \includegraphics[width=\textwidth, height=10cm]{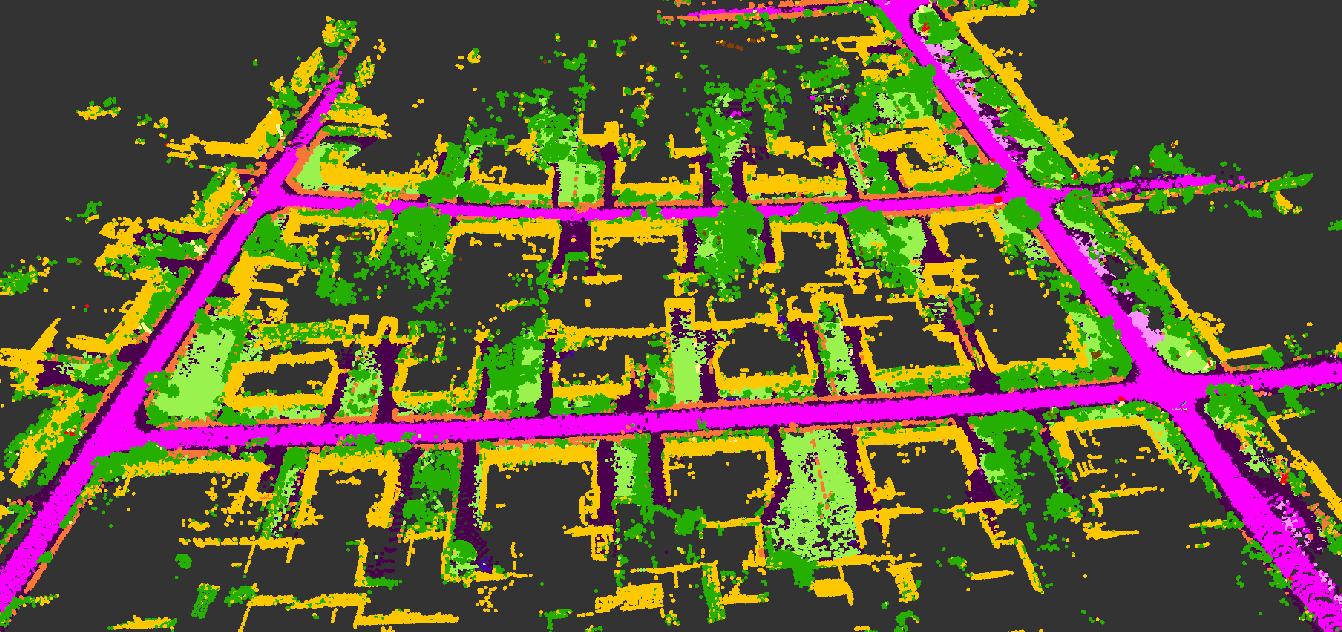}
\end{subfigure}

\caption{Samples of the output semantic maps built using our framework}
\label{fig:semantic_maps}
\end{figure}

Figure \ref{fig:trajectories} shows the trajectories of different methods on the odd KITTI sequences when skipping 0, 5, and 10 scans between each two processed scans, respectively. In contrast to SUMA++, our approach uses semantic information to improve the matching process itself, leading to better results for fast moving platforms, while SUMA++ uses semantic information only to assign weights to matches during the optimization process. As we drop more frames, trajectories corresponding to our framework with improving matches using both semantic information and outliers rejection are shown to be usually closer to the ground truth trajectories than other methods. Figure \ref{fig:semantic_maps} shows samples of the output semantic maps constructed using our framework for KITTI sequence 05. The deep semantic segmentation model has never seen the LiDAR scans of this sequence during training.

\begin{table*}[!htbp]
\caption{Sequence-wise trajectory percentage error for KITTI odd sequences}
\label{tab:all_sequences}

\begin{center}

\begin{subtable}{\textwidth}
\centering
\begin{tabular}{l|c|c|c|c|c|c}
\hline
Method / Sequence & 01 & 03 & 05 & 07 & 09 & Average \\
\hline

No semantics, no outliers rejection & 2.52 & 1.12 & 0.72 & 0.64 & 1.09 & 1.22 \\
\hline

Using semantics, no outliers rejection & 2.71 & 1.08 & 0.79 & 0.63 & 1.16 & 1.27 \\
\hline

Using semantics, using outliers rejection & 2.71 & \textbf{1.07} & 0.73 & 0.64 & 1.16 & 1.26 \\
\hline

SUMA++ & \textbf{1.71} & 1.74 & \textbf{0.48} & \textbf{0.37} & \textbf{0.54} & \textbf{0.97}\\
\hline

\end{tabular}
\caption{Number of skipped scans = 0}
\end{subtable}

\texttt{\\}
\texttt{\\}

\begin{subtable}{\textwidth}
\centering
\begin{tabular}{l|c|c|c|c|c|c}
\hline
Method / Sequence & 01 & 03 & 05 & 07 & 09 & Average \\
\hline

No semantics, no outliers rejection & 2.59 & 1.07 & 0.76 & 0.64 & 1.18 & 1.25 \\
\hline

Using semantics, no outliers rejection & 2.86 & \textbf{1.05} & 0.73 & 0.67 & 1.24 & 1.31 \\
\hline

Using semantics, using outliers rejection & 2.85 & \textbf{1.05} & 0.74 & 0.67 & 1.23 & 1.31 \\
\hline

SUMA++ & \textbf{2.12} & 1.79 & \textbf{0.61} & \textbf{0.45} & \textbf{0.61} & \textbf{1.12}\\
\hline

\end{tabular}
\caption{Number of skipped scans = 1}
\end{subtable}

\texttt{\\}
\texttt{\\}
















\begin{subtable}{\textwidth}
\centering
\begin{tabular}{l|c|c|c|c|c|c}
\hline
Method / Sequence & 01 & 03 & 05 & 07 & 09 & Average \\
\hline

No semantics, no outliers rejection & \textbf{2.61} & 1.51 & 0.85 & 0.69 & \textbf{1.29} & 1.39 \\
\hline

Using semantics, no outliers rejection & 2.83 & \textbf{1.05} & 0.86 & 0.73 & 1.42 & 1.38 \\
\hline

Using semantics, using outliers rejection & 2.82 & 1.06 & 0.88 & 0.69 & 1.41 & \textbf{1.37} \\
\hline

SUMA++ & 7.15 & 1.87 & \textbf{0.75} & \textbf{0.68} & 5.05 & 3.10\\
\hline

\end{tabular}
\caption{Number of skipped scans = 4}
\end{subtable}

\texttt{\\}
\texttt{\\}

\begin{subtable}{\textwidth}
\centering
\begin{tabular}{l|c|c|c|c|c|c}
\hline
Method / Sequence & 01 & 03 & 05 & 07 & 09 & Average \\
\hline

No semantics, no outliers rejection & 6.75 & 3.39 & 4.02 & 2.98 & \textbf{1.34} & 3.70 \\
\hline

Using semantics, no outliers rejection & 2.88 & 2.66 & 1.01 & 0.85 & \textbf{1.34} & 1.75 \\
\hline

Using semantics, using outliers rejection & \textbf{2.63} & \textbf{1.02} & \textbf{0.93} & \textbf{0.78} & 1.44 & \textbf{1.36} \\
\hline

SUMA++ & 26.55 & 4.23 & 39.65 & 1.02 & 8.53 & 15.99\\
\hline

\end{tabular}
\caption{Number of skipped scans = 8}
\end{subtable}

\texttt{\\}
\texttt{\\}

\begin{subtable}{\textwidth}
\centering
\begin{tabular}{l|c|c|c|c|c|c}
\hline
Method / Sequence & 01 & 03 & 05 & 07 & 09 & Average \\
\hline

No semantics, no outliers rejection & 6.78 & 23.61 & 4.00 & 2.48 & \textbf{1.33} & 7.64 \\
\hline

Using semantics, no outliers rejection & 3.50 & 2.29 & \textbf{0.82} & \textbf{0.73} & 1.44 & 1.76 \\
\hline

Using semantics, using outliers rejection & \textbf{2.83} & \textbf{1.02} & 0.89 & 0.74 & 1.48 & \textbf{1.39} \\
\hline

SUMA++ & 17.38 & 4.81 & 5.56 & 5.18 & 9.79 & 8.54\\
\hline

\end{tabular}
\caption{Number of skipped scans = 9}
\end{subtable}

\texttt{\\}
\texttt{\\}
\begin{subtable}{\textwidth}
\centering
\begin{tabular}{l|c|c|c|c|c|c}
\hline
Method / Sequence & 01 & 03 & 05 & 07 & 09 & Average \\
\hline

No semantics, no outliers rejection & 93.38 & 22.60 & 7.46 & 16.19 & 4.82 & 28.89 \\
\hline

Using semantics, no outliers rejection & 12.33 & 2.80 & \textbf{0.85} & \textbf{0.70} & 1.44 & 3.62 \\
\hline

Using semantics, using outliers rejection & \textbf{2.61} & \textbf{0.95} & 0.91 & 0.71 & \textbf{1.39} & \textbf{1.32} \\
\hline

SUMA++ & 53.84 & 2.32 & 5.95 & 26.35 & 14.46 & 20.58\\
\hline

\end{tabular}
\caption{Number of skipped scans = 10}
\end{subtable}

\end{center}

\end{table*}

\subsubsection{Quantitative Evaluation}
\label{sec:quantitative}
Trajectory estimation errors are reported in Table \ref{tab:all_sequences}. These are the average percentage errors as explained in Section \ref{sec:dataset}. Although we expect the error to get monotonically worse as we drop more frames, we observe that the error behaviour is not monotonic for some sequences. The justification for this is that as we adjust the system parameters according to the number of skipped scans as we explained above, the adequacy of the system parameters to the operation conditions can vary from one sequence to another. Automatic tuning of such parameters can be an interesting research point for future work.

\begin{figure}[tb]
\centering
  \includegraphics[width=1\linewidth]{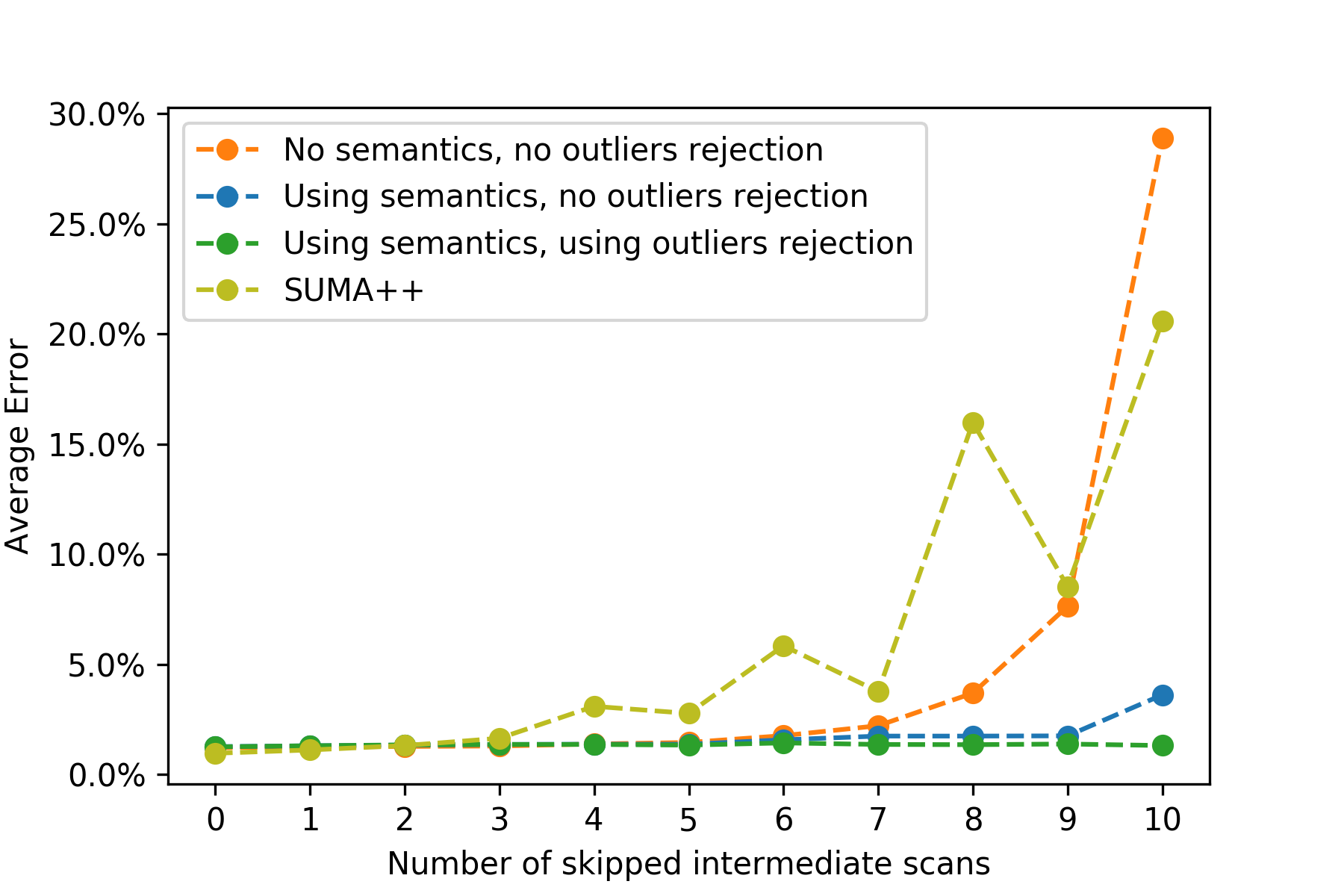}
  \caption{Average percentage error over KITTI odd sequences for different numbers of skipped scans}
  \label{figure:error_curve}
\end{figure}

To observe the overall trend, Figure \ref{figure:error_curve} shows the average error over the odd sequences for different numbers of skipped scans, and we take the macro average over all odd sequences. The figure shows that when dropping no or few scans between each two processed scans (low speed motion), improving matching using semantic information and outliers rejection does not improve the total error, and it may even make the error worse in some scenarios due to the inherent errors in the semantic segmentation and outliers rejection processes. This is because when we drop no frames, the relative translation and rotation between two consecutive scans is small, and therefore most matches that we get using nearest neighbors search are correct matches that are good enough to estimate the motion transformation as we discussed in Section \ref{sec:qualitative}. However, when we drop more scans (which simulates the case where the platform moves at a higher speed), improving matches via exploiting semantic information and our novel outliers rejection method is shown to be of significant importance as demonstrated by Figure \ref{figure:error_curve}.

\begin{figure}[tb]
\centering
  \includegraphics[width=1\linewidth]{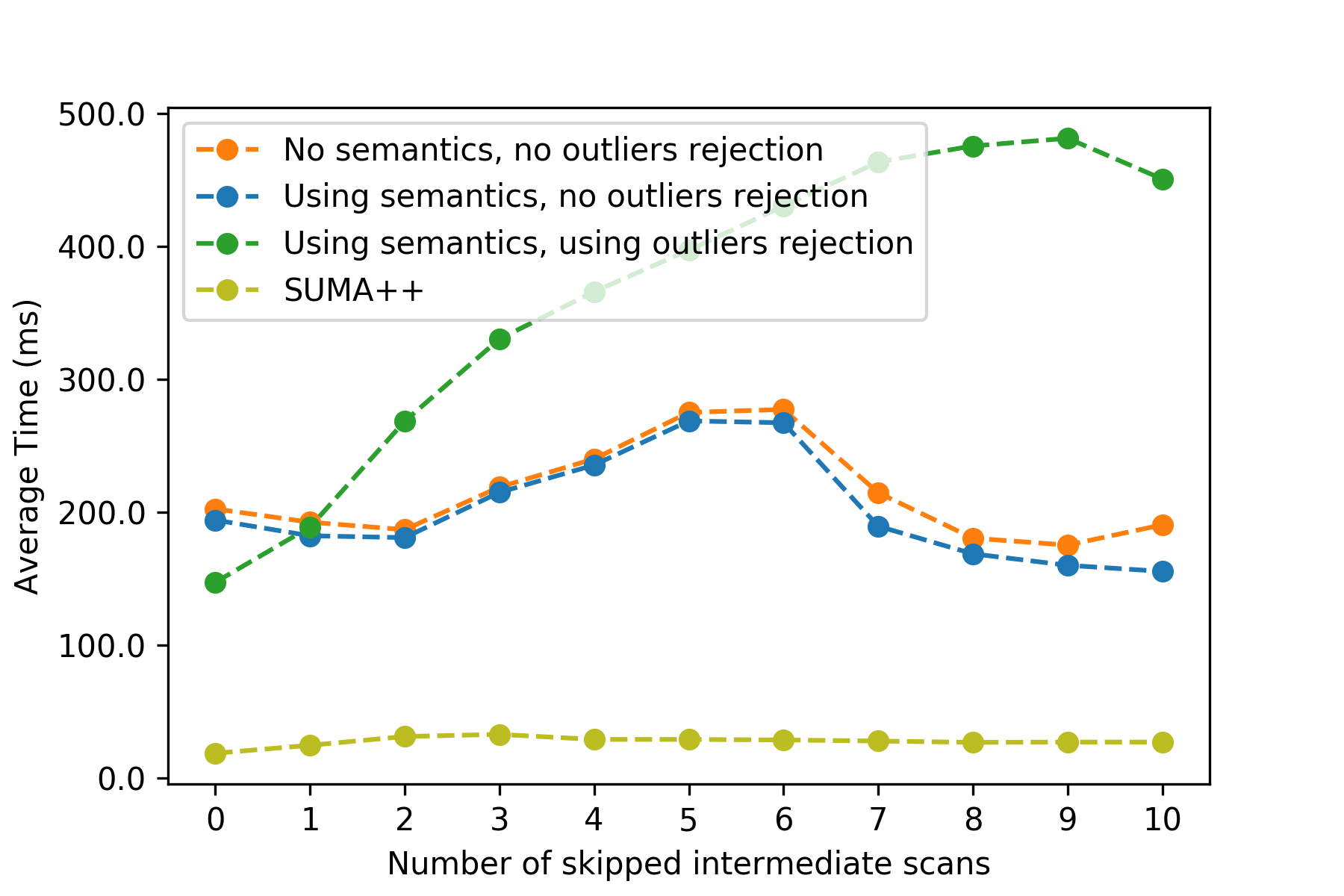}
  \caption{Average time over odd sequences for different numbers of skipped scans}
  \label{figure:time_curve}
\end{figure}

\subsubsection{Processing Time}
\label{sec:processing_time}
We report the average time used to process a scan received from the LiDAR. All experiments were conducted on an Intel i7-9700K CPU. The time consumed by the deep semantic segmentation model \cite{DBLP:journals/corr/abs-2011-01974} is not included, as it is supposed to be running on GPU in parallel to the LiDAR Odometry and Mapping module. We use a C++ implementation with Ceres Solver \cite{ceres-solver} as the optimization backend. Open Multi-Processing \footnote{\url{https://en.wikipedia.org/wiki/OpenMP}} is used to parallelize some of the operations when possible using 8 threads.

Figure \ref{figure:time_curve} shows the average scan processing time over all odd sequences against the number of dropped scans. When using outliers rejection, we terminate scan-to-scan matching if algorithm \ref{alg:orme} converges from the second iteration, which indicates the detection of no or very few outlier matches. This trick makes the average scan processing time of our framework even less than when not using outliers rejection by $25\%$ when dropping no scans. As we drop more intermediate scans, the framework consumes more time as needed. The figure also shows that exploiting the semantic information consistently improves processing time over the case when the semantic information is not used. We can also observe that when we drop more scans, the average processing time starts decreasing after some point. This is because the large gap between the two processed scans will lead to low overlap and less matches during registration, which will in turn cause the scan matching time to decrease. We also report the average processing time for SUMA++, which is significantly faster because SUMA++ applies scan matching on spherical projection images instead of matching raw 3D point clouds. While LOAM reported processing speed for KITTI LiDAR scans is 1 scan per second \cite{zhang2017low}, our framework speed is about 6.67 scans per second. Real-time processing of KITTI scans (10 scans per second) can be achieved by skipping a single intermediate frame, at the cost of increasing the average error from 1.26\% to 1.31\%.

In another experiment, we run the framework with 10 times the number of maximum allowed ICP iterations without outliers rejection. The overall average error when dropping 10 scans is 1.84\% and the average processing time is 484 ms. Using outliers rejection improves both the error and processing time to 1.32\% and 445 ms, which proves that our outliers rejection algorithm has an added value over simply doing more ICP iterations with robust M-estimation techniques and re-computing the matches using nearest neighbors search after each iteration.

\section{Conclusions and Future work} \label{sec:future}

\subsection{Conclusion}

In this work, we presented a robust framework for real-time LiDAR odometry and mapping for fast moving platforms. The pipeline starts by passing the input LiDAR scan through a deep semantic segmentation model that we proposed and trained to achieve high accuracy and real time segmentation of LiDAR point clouds. 

We investigated the exploitation of semantic information during the step of keypoints matching, and showed through a set of experiments on KITTI dataset that this has a major impact on localization accuracy in the case of large gaps between LiDAR scans acquisition poses (which would arise when the platform moves at high speed).

We also investigated the matching error patterns that arise when we select matches according to semantic information, and designed a special outliers detection and rejection techniques to further improve the framework robustness. 

\subsection{Future Directions}
The proposed framework for robust LiDAR odometry and mapping still has room for improvement. The main directions for improving the framework are:
\begin{itemize}
    \item As we mentioned in Section \ref{sec:results}, most SLAM frameworks have many parameters that require custom manual tuning for each environment and operating conditions. These parameters should also be tuned according to the expected vehicle speed. One way to improve our framework is to add a module that estimates the vehicle speed and automatically decides the values of the different parameters accordingly. 
    
    \item We can employ LiDAR scan motion distortion correction techniques to allow using the framework with datasets that do not preprocess the LiDAR scans and provide corrected points clouds. We can then use simulations like Carla car simulator \cite{Dosovitskiy17} or Airsim drone simulator \cite{airsim2017fsr} to simulate vehicles moving at different speeds in the presence of motion distortion to test our framework at different speeds in a more realistic setup. We can also experiment on other datasets such as KITTI-360 \cite{abu2018augmented} which contains real images augmented with virtual objects, or create our own real-world dataset by operating and collecting LiDAR data mounted on a drone or car.
    
    \item Investigate which object classes are more important in motion estimation, and give more emphasis to these classes during semantic segmentation training. 

    \item Study the impact of semantic segmentation errors on the quality of localization and mapping.

    \item Use loop closure detection and correction to generate globally consistent trajectory estimations, while exploiting semantic information in loop closure detection.
    
    \item Optimize the framework processing time to enable real-time processing on low-end embedded devices.
\end{itemize}

\vspace{6pt} 

\begin{adjustwidth}{-\extralength}{0cm}

\reftitle{References}

\bibliography{mybibfile.bib}

%


\end{adjustwidth}
\end{document}